\PassOptionsToPackage{table}{xcolor}
\documentclass{style}
\microtypesetup{expansion=false}
\usepackage[utf8]{inputenc}
\usepackage{amsmath,amsfonts,amssymb}
\usepackage{amsmath,amsfonts,bm}

\def\eqref#1{equation~\ref{#1}}
\def\1{\bm{1}}

\DeclareMathAlphabet{\mathsfit}{\encodingdefault}{\sfdefault}{m}{sl}
\SetMathAlphabet{\mathsfit}{bold}{\encodingdefault}{\sfdefault}{bx}{n}

\usepackage{nicefrac,enumitem,wrapfig,pifont}
\newcommand{\cmark}{\textcolor{green}{\ding{51}}}
\newcommand{\xmark}{\textcolor{red}{\ding{55}}}
\newcommand{\name}{VideoZeroBench}
\hypersetup{
  pdftitle={VideoZeroBench: Probing the Limits of Video MLLMs with Spatio-Temporal Evidence Verification},
  pdfauthor={Jiahao Meng, Yue Tan, Qi Xu, Haochen Wang, Zhongwei Ren, Weisong Liu, Yuhao Wang, Renrui Zhang, Xiangtai Li, Haodong Duan, Yunhai Tong, Ming-Hsuan Yang}
}
\title{VideoZeroBench: Probing the Limits of Video MLLMs with Spatio-Temporal Evidence Verification}
\author[1]{Jiahao Meng}
\author[1]{Yue Tan}
\author[2]{Qi Xu}
\author[3]{Haochen Wang}
\author[4]{Zhongwei Ren}
\author[3]{Weisong Liu}
\author[1]{Yuhao Wang}
\author[1]{Ray Zhang}
\author[1]{Jason Li}
\author[1]{Kenny Duan}
\author[1,\dagger]{Yunhai Tong}
\author[5]{\\ \mbox{Ming-Hsuan Yang}}
\affiliation[]{PKU$^1$ \quad WHU$^2$ \quad CASIA$^3$ \quad BJTU$^4$ \quad UC Merced$^5$}
\contribution[\dagger]{Corresponding Author}
\checkdata[Project Page]{\url{https://marinero4972.github.io/projects/VideoZeroBench}}

\abstract{
Video multimodal large language models achieve strong results on existing benchmarks, but answer accuracy alone does not establish whether they can locate the evidence needed to answer a question.
We introduce \textbf{\name{}}, a challenging long-video benchmark with manually annotated question--answer pairs spanning 13 video domains.
Questions target fine-grained cues, fleeting events, and evidence distributed across multiple segments. Temporal intervals and key-frame boxes are annotated where applicable.
All questions undergo two rounds of cross-verification for answer validity and evidence quality.
Our \textbf{five-level diagnostic protocol} compares answering with and without evidence hints, then combines answer correctness with independently evaluated temporal and spatial grounding.
Across 19 evaluated models, the best standard QA accuracy is 24.8\% (Level-3), achieved by Gemini-3.7-Flash. No model exceeds 1.8\% when correct answers and accurate spatio-temporal localization are jointly required (Level-5).
Analyses of atomic abilities, evidence spans, input modalities, and thinking-with-videos inference further characterize where the evaluated systems struggle.
These findings motivate more precise evidence search and localization for long-video question answering.
Our code and data are publicly released.
}

\begin{document}
\maketitle

\begin{figure}[t]
\centering

\includegraphics[width=\linewidth]{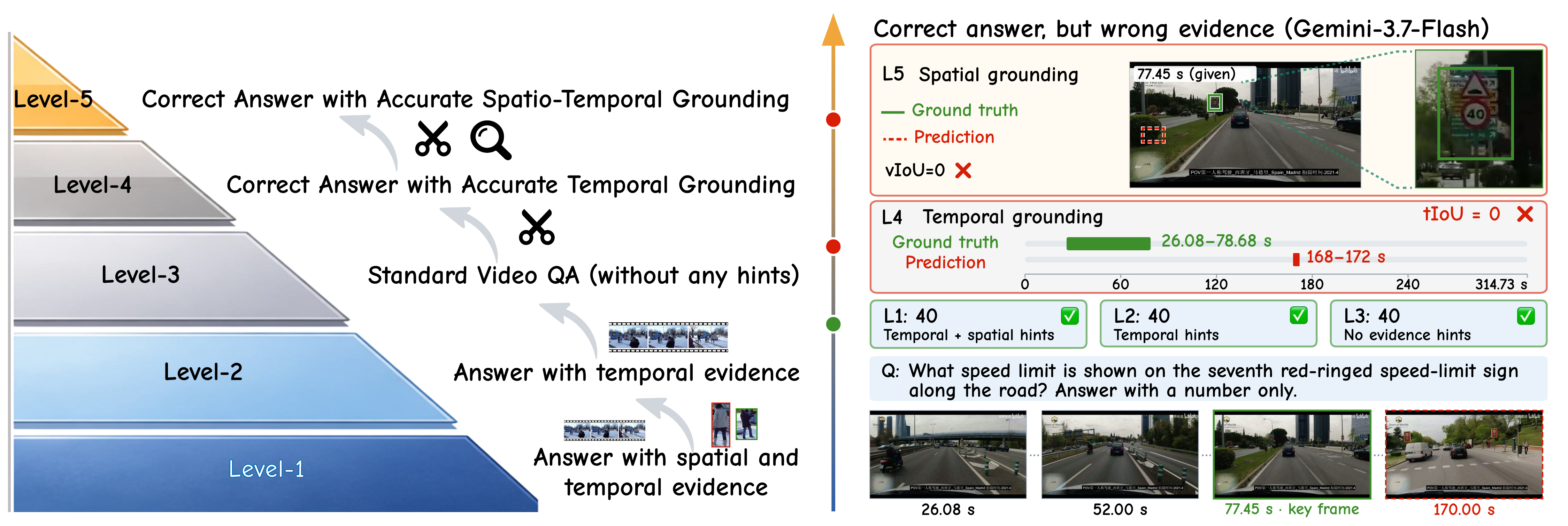}
\caption{
\small We introduce \textbf{\name{}}, a new challenging long-video understanding benchmark with hierarchical spatio-temporal evidence verification.
The example highlights a reliability gap: even the frontier model Gemini-3.7-Flash answers correctly but may still fail to provide accurate spatio-temporal evidence.
}
\label{fig:teaser}
\end{figure}

\section{Introduction}
\label{sec:intro}

%%% SOTA through 2026-08-31; provenance and protocol checks: ../SOTA_SOURCES_2026-08.md
% 参考：
% https://huggingface.co/Qwen/Qwen3.5-397B-A17B
% https://seed.bytedance.com/en/seed2
% https://qwen.ai/blog?id=qwen3.8 (2026-08-03, multimodal evaluation table)
% https://huggingface.co/moonshotai/Kimi-K3/blob/a590ce090cb049c93a33dfe8c208ec652aa20503/README.md
% https://seed.bytedance.com/en/seed2_1 (Model Card, Tables 9--10)
% https://github.com/zerobench/zerobench.github.io/blob/1b05f0a4c91460b33700be01bf2d4e72d5e17d3b/index.html

Multimodal large language models (MLLMs)~\citep{gemini-37-flash,gpt-5.6-sol} have made progress in video understanding, scoring over 80\% on Video-MME~\citep{VideoMME} and LongVideoBench~\citep{LongVideoBench} (Table~\ref{tab:bench_comparison}).
However, high QA accuracy can \textbf{obscure capability boundaries} in fine-grained perception and long-range reasoning, especially when supporting evidence is not verified.
Two concerns remain: \textbf{capability}, whether models can locate critical visual details and reason over them, and \textbf{reliability}, whether their answers are supported by correct spatio-temporal evidence.
Progress requires evaluating both answer correctness and evidence grounding.

Several recent video benchmarks, such as V-STaR~\citep{VSTaR}, ToG-Bench~\citep{xu2025tog}, Know-Show~\citep{sugandhika2025know}, and Dr.V~\citep{luo2025dr}, incorporate temporal grounding and spatial grounding as additional evaluation dimensions.
However, VideoZeroBench differs from these benchmarks in three aspects.
First, \name{} decomposes difficult video QA into temporal evidence search, spatial evidence search, and answer reasoning through its hierarchical design.
Second, \name{} emphasizes fine-grained cues such as fleeting frames, small objects, subtle spatial relations, and non-salient evidence.
Third, its videos are longer on average than those in the compared spatio-temporal grounding benchmarks (Table~\ref{tab:bench_comparison}), increasing the temporal search space.
Detailed comparisons and representative cases are provided in Appendix~\ref{sec:ap_case_comparison} and Figure~\ref{fig:ap_example_comparison}.
Meanwhile, a growing number of thinking-with-videos approaches~\citep{zeng2026videoo3,yan2025videochat,meng2025open,ding2025videozoomer} adopt iterative evidence retrieval and reasoning.
These approaches interleave reasoning with explicit visual grounding to seek key frames, timestamps, or spatial regions.
However, existing benchmarks provide limited verification of their spatio-temporal evidence on difficult video QA.
Overall, unlike the image domain, which has benchmarks such as ZeroBench~\citep{roberts2025zerobench} and TreeBench~\citep{wang2025traceable}, the video domain still lacks a sufficiently challenging benchmark that probes fine-grained spatio-temporal reasoning and verifies evidence grounding.
Further discussion of related work is provided in Appendix~\ref{sec:ap_related_works}.

To address these limitations, we introduce \textbf{\name{}}, a capability-centric, evidence-grounded, and hierarchical benchmark for video understanding.
It aims to explore the limits of leading video MLLMs by examining shared weaknesses in answering challenging questions and grounding answers in spatio-temporal evidence.
\name{} contains manually annotated, open-ended questions across 13 video domains, including sports, instructional videos, life vlogs, and games.
Its videos average 667.1 seconds, introducing complex scenes and long-range temporal dependencies.
%
% REVIEW[R06]: Check whether multi-segment causal dependency and multi-segment dependency denote the same category.
The benchmark covers 11 atomic abilities, including counting, small-object perception, spatial orientation discrimination, and multi-segment dependency, with many questions requiring multiple abilities.
All questions require precise and verifiable answers (e.g., numbers, single words, or fixed short phrases), enabling direct answer comparison without an LLM judge.
In summary, \textbf{\name{} has three key features: }
\textbf{(1) High difficulty.} Questions are manually annotated and intentionally challenging, designed to expose the capability boundaries of contemporary video MLLMs under fine-grained and complex spatio-temporal conditions.
\textbf{(2) Evidence-grounded evaluation.} We evaluate answer correctness together with temporal and spatial evidence localization on the same questions.
\textbf{(3) Manually verified annotations.} PhD-level annotators construct the questions and evidence. Two-round cross-verification checks answer uniqueness and evidence completeness.

\begin{table}[t]
\centering
\small
\setlength{\tabcolsep}{3pt}
\renewcommand{\arraystretch}{0.90}
\caption{\small Comparison with video and spatial benchmarks. TG/SG: temporal/spatial grounding; MCQ/OE: multiple-choice/open-ended; Anno: annotation source (M: manual, A: automatic). Scores use each benchmark's native protocol and are not directly comparable. The \name{} QA count denotes distinct questions, and its score is Level-5 accuracy.}
\scalebox{1.0}{
\begin{tabular}{lccccccccc}
\toprule
\textbf{Benchmark} & \textbf{Videos} & \textbf{Duration} & \textbf{QAs} & \textbf{Type} & \textbf{Anno}& \textbf{TG} & \textbf{SG} & \textbf{Hierarchic} & \textbf{SOTA(2026.08)} \\
\midrule

\multicolumn{10}{l}{\textit{\textbf{I. General Video MLLM Benchmarks}}} \\
\midrule
Video-MME~\citep{VideoMME} & 900 & 1017.9s & 2.7K & MCQ & M &\xmark & \xmark & \xmark & 90.4\% \\
MVBench~\citep{MVBench} & 3.6K & 16s & 4K & MCQ & A\&M & \xmark & \xmark & \xmark & 78.1\% \\
Video-MMMU~\citep{VideoMMMU} & 300 & 506.2s & 900 & MCQ/OE & M & \xmark & \xmark & \cmark & 88.7\%\\
LongVideoBench~\citep{LongVideoBench} & 3.8K & 473s & 6.6K & MCQ & M &\xmark & \xmark & \cmark &  80.6\% \\
CG-Bench~\citep{CGBench} & 1.2K & 1624.4s & 12K & MCQ/OE & M & \cmark & \xmark & \xmark & 65.5\%  \\

\midrule
\multicolumn{10}{l}{\textit{\textbf{II. Detailed Spatial Benchmarks}}} \\
\midrule
ZeroBench~\citep{roberts2025zerobench}~(Image) & - & - & 100 & OE & M & \xmark & \xmark & \xmark & 30\% \\
BLINK~\citep{fu2024blink}~(Image)  & - & - & 3.8K & MCQ & M &\xmark & \xmark & \xmark & 81.4\% \\
TreeBench~\citep{wang2025traceable}~(Image) & - & - & 405 & MCQ & A\&M  &\xmark & \cmark & \xmark &  71.1\% \\
% Spatial-Tree & \multicolumn{2}{c}{[Image and Video]}& 6.3K & MCQ\&OE & A\&M &\xmark & \xmark & \cmark & 57.8\% \\
MMSI-Video-Bench~\citep{lin2025mmsi} & 1.3K & 72s & 1.1K & MCQ & M &\xmark & \xmark & \cmark & 38.0\% \\
V-STaR~\citep{VSTaR} & 743 & 110.23s & 2.1K & OE & A\&M & \cmark & \cmark & \xmark & 46.6\%\\
% Know-Show & 750 &  & 2.5K & OE & M & \cmark & \cmark & \xmark & \\
ToG-Bench~\citep{xu2025tog} & 100 & 87.9s & 2.7K & OE & A\&M &\cmark & \cmark & \xmark & 89.4\% \\
\midrule
\name{} & 138 & 667.1s & 500 & OE & M &\cmark & \cmark & \cmark & 1.8\% (Level-5) \\
\bottomrule
\end{tabular}}
\label{tab:bench_comparison}
\end{table}

To distinguish answer correctness from correct evidence grounding, we also introduce a \textbf{five-level} hierarchical evaluation framework (Figure~\ref{fig:teaser}).
The first three levels evaluate answering under different evidence conditions: \textbf{Level-1} provides both temporal and spatial evidence, \textbf{Level-2} provides temporal evidence only, and \textbf{Level-3} corresponds to standard video QA without hints.
The remaining levels require grounding: \textbf{Level-4} requires correct answers with accurate temporal segments, while \textbf{Level-5} further requires both temporal and spatial localization.
The five levels reuse the same question set, with evidence annotations available where applicable (Section~\ref{sec:stats}).
This hierarchy compares answering with and without evidence hints and evaluates temporal and spatial predictions for the same questions.
It reveals localization weaknesses that answer-only scores cannot distinguish.
In Figure~\ref{fig:teaser}, Gemini-3.7-Flash answers correctly at Levels 1--3 but fails both temporal and spatial grounding.

Experiments show that answering these questions correctly often does not coincide with accurate spatio-temporal localization.
Gemini-3.7-Flash achieves the highest Level-3 accuracy at 24.8\%, followed by GPT-5.6-Sol at 18.8\%.
\textbf{Under the strictest Level-5 requirement, which demands both correct answers and correct spatio-temporal grounding, all models score at most 1.8\%, with many achieving zero.}
Independent human participants reach 63.5\% and 55.8\% at Levels 4 and 5 on a 50-question subset (Appendix~\ref{sec:ap_human}).

We further analyze performance variations and error cases across video categories, atomic abilities, evidence span lengths, thinking-with-videos inference, and input modalities, leading to three conclusions.
\textbf{First, models may answer challenging questions correctly yet fail to localize the supporting spatio-temporal evidence}.
Incorrect grounding predictions remain common (Table~\ref{tab:main_results}, Figures~\ref{fig:teaser} and~\ref{fig:cases}).
\textbf{Second, fine-grained spatial perception and temporal evidence search remain challenging} (Figure~\ref{fig:analysis}).
\textbf{Third, thinking-with-videos inference} improves QA in the VideoChat-R1.5 ablation, but \textbf{does not improve grounding accuracy in this setting} (Table~\ref{tab:agentic}).
These findings highlight the need to prioritize evidence-grounded perception and precise spatio-temporal reasoning as foundations of trustworthy video intelligence.
% %
We hope that \name{} can serve as a rigorous benchmark for diagnosing and advancing the next generation of video MLLMs toward reliable and verifiable understanding.

\section{\name{}}
\label{sec:method}

\subsection{Dataset Construction}

\begin{figure}[t]
    \centering
    \includegraphics[width=1.0\linewidth]{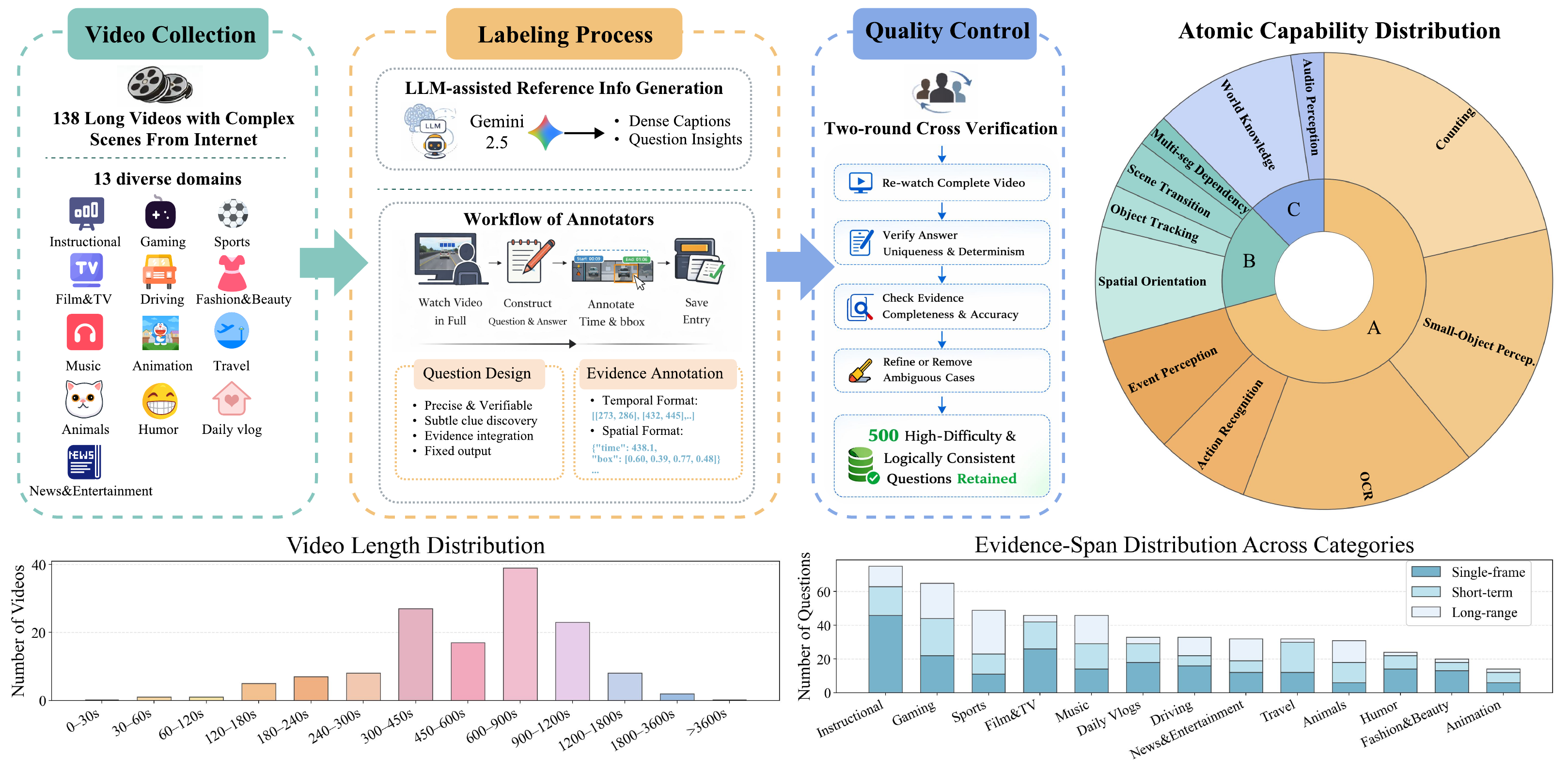}
    \caption{
\small Data construction and statistics of \name{}.
All questions and evidence are manually annotated and verified. The benchmark spans 13 video domains and covers 11 atomic capabilities grouped into Detailed Perception (A), Spatial\&Temporal Reasoning (B), and Semantic\&Cross-Modal Reasoning (C). The bottom plots show the distributions of video length and minimal evidence span across categories.
}
    \label{fig:data}
\end{figure}

\noindent
\textbf{Video Collection.}
We manually curate long-duration, structurally complex videos from YouTube and Bilibili, without using existing benchmark datasets.
Annotators search by category and manually review candidates, prioritizing uploads from the preceding six months where possible.
Videos are included only if they exhibit meaningful event progression, non-trivial scene transitions, and sufficient visual complexity such that critical evidence is not always salient.
Short or visually simple clips are excluded.
Most retained videos range from 5 to 20 minutes, increasing temporal depth and dispersing supporting evidence across time.
As shown in Figure~\ref{fig:data}, the collection spans 13 domains, including Instructional, Gaming, Sports, Film\&TV, Music, Daily Vlogs, Driving, News\&Entertainment, Travel, Animals, Humor, Fashion\&Beauty, and Animation.
Both Chinese and English videos are incorporated to enhance linguistic diversity.
The resulting benchmark comprises long, cluttered videos in which key evidence is subtle and temporally distributed.

\noindent
\textbf{Labeling Process.}
The annotation pipeline includes LLM-assisted reference generation followed by fully manual question and evidence construction.
We first apply Gemini-2.5-Pro~\citep{gemini2.5} to produce dense captions and candidate question insights for each video.
These outputs serve only as auxiliary references to help annotators understand long videos and expand the question design space, and they do not determine ground-truth labels.
Annotators then watch each video in full and manually construct questions with precise and verifiable answers.
All questions follow a fixed-output format, such as numbers, single words, or short phrases.
Questions that can be answered by simple event recognition or straightforward object identification are avoided.
Instead, annotators design problems that require subtle clue discovery or integration of non-salient evidence across frames.

For each question, temporal intervals are annotated when the question is explicitly associated with one or multiple distinct time spans in the video.
Bounding boxes are annotated on key frames when the question clearly depends on specific small objects or spatial regions within those frames.
Temporal evidence is stored as structured start–end timestamps, and spatial evidence is recorded as normalized bounding boxes associated with timestamps.
This explicit representation enables systematic verification and quantitative evaluation of spatio-temporal grounding.

\noindent
\textbf{Quality Control.}
All questions undergo two-round cross-verification by annotators within the same team, in which each question is reviewed by members who do not participate in its original annotation.
Reviewers rewatch the complete video to ensure that each question is fully answerable using the annotated evidence.
They verify the uniqueness and determinism of answers under the specified format.
Ambiguous, under-specified, or logically flawed questions are revised or removed.
Evidence annotations are refined if incomplete or redundant.
Atomic ability categories and minimal evidence span labels are also confirmed during verification.
This targeted diagnostic set prioritizes detailed evidence annotation and repeated verification over maximizing question count, supporting focused analysis of answering and grounding failures. Annotation and filtering details are provided in Appendix~\ref{sec:ap_data}.

\subsection{Benchmark Statistics}
\label{sec:stats}

\name{} contains 500 distinct question--answer pairs from 138 manually curated videos across 13 domains, with 280 Chinese and 220 English questions.
Of these questions, 442 have temporal intervals, 372 have key-frame boxes, and 320 have both. Appendix~\ref{sec:ap_exp} details annotation coverage.
Figure~\ref{fig:data} summarizes video categories, capabilities, durations, and evidence spans.
Videos total 25.57 hours and range from 30 to 3,037 seconds.
Their average duration is 667.1 seconds, and most last 5--20 minutes.
Compared with recent grounding benchmarks~\citep{VSTaR,xu2025tog}, the videos are longer on average (Table~\ref{tab:bench_comparison}), requiring search over longer timelines.
Annotated boxes cover 6.8\% of the frame area on average.
Fine-grained evidence includes small objects, spatial relations, fleeting actions, and object tracks.

Error patterns observed during annotation and cross-verification motivate a taxonomy of 11 atomic capabilities.
\textbf{Detailed Perception} includes counting, small-object perception, OCR, action recognition, and event perception.
These abilities test fine-grained visual perception in cluttered scenes.
\textbf{Spatial \& Temporal Reasoning} includes spatial orientation discrimination, object tracking, scene transition understanding, and multi-segment dependency.
These abilities test how models integrate spatial relations and temporal cues.
\textbf{Semantic \& Cross-Modal Reasoning} includes world knowledge reasoning and audio perception.
These abilities require combining visual evidence with contextual knowledge or auditory cues.
This taxonomy supports the capability-level analysis in Figure~\ref{fig:analysis}.

We also group questions by their \textbf{minimal evidence span}: the shortest temporal coverage needed to answer them (Figure~\ref{fig:data}).
The 216 single-frame questions need only one frame.
The 155 short-term questions require continuous observation over less than 15 seconds.
The 129 long-range questions require extended observation or evidence integration across temporally distant segments.
This categorization describes evidence distribution and the information needed for reasoning.

\subsection{Five-Level View of \name{}}
\label{sec:fivelevel}

The five-level hierarchy diagnoses answering and evidence localization under different input conditions.
Levels 1--3 compare QA with progressively fewer evidence hints. Levels 4--5 combine Level-3 answer correctness with independently prompted temporal and spatial predictions.
The resulting scores measure success across tasks for the same question.

\noindent
\textbf{Level-1: Answer with Spatial and Temporal Evidence.}
In Level-1, the available temporal intervals and key spatial regions are provided in textual format, along with the entire video.
The performance is evaluated using standard QA accuracy.
Errors at this level may reflect difficulties in perceiving, integrating, or reasoning over the indicated evidence.

\noindent
\textbf{Level-2: Answer with Temporal Evidence.}
The input includes a query, a full video, and the available temporal annotations, without spatial region hints.
By restricting the evidence to temporal cues only, this setting makes questions that depend on fine-grained spatial perception or small-object localization more difficult.
Performance is measured by QA accuracy.

\noindent
\textbf{Level-3: Answer without Evidence Hints.}
The model receives only the video and the question.
This is the standard end-to-end QA setting, and Level-3 performance is evaluated based on QA accuracy.
Correct answers here are not required to be grounded in verified evidence.

\noindent
\textbf{Level-4: Correct Answer with Accurate Temporal Grounding.}
The model must both answer the question correctly and identify one or more supporting temporal intervals.
At this level, we feed the original video and the question in a separate prompt and require the model to output only the key time intervals.
We first compute multi-segment temporal IoU (tIoU).
All five accuracy scores use the same denominator $N=500$.
Let $\mathcal{T}$ denote the 442 questions with temporal annotations.
We combine temporal grounding with Level-3 answer correctness to compute the final Level-4 accuracy as follows:
\begin{equation}
\mathrm{Accuracy}_{L4}
=
\frac{1}{N}
\sum_{i\in\mathcal{T}}
\mathbb{I}(\hat{y}_i = y_i)
\cdot
\mathbb{I}(\mathrm{tIoU}_i > 0.3).
\end{equation}
where $y_i$ is the ground-truth answer and $\hat{y}_i$ denotes the model's predicted answer from Level-3. A question passes when its answer and independently predicted temporal evidence are both correct.

\noindent
\textbf{Level-5: Correct Answer with Accurate Spatio-Temporal Grounding.}
The model must correctly answer the question, identify temporal intervals, and accurately localize objects with bounding boxes at annotated timestamps.
At this level, we input the original video, the question, and the key-frame timestamps and require the model to output 2D boxes at those timestamps.
Fixed timestamps allow comparison with sparse manual boxes at the same frames despite object motion.
For scoring, we first compute the visual IoU for one or more boxes in each frame, then average these scores over frames to obtain vIoU.
Let $\mathcal{S}$ denote the 372 questions with spatial annotations.
The final Level-5 accuracy is:
\begin{equation}
\mathrm{Accuracy}_{L5}
=
\frac{1}{N}
\sum_{i\in\mathcal{T}\cap\mathcal{S}}
\mathbb{I}(\hat{y}_i = y_i)
\cdot
\mathbb{I}(\mathrm{tIoU}_i > 0.3)
\cdot
\mathbb{I}(\mathrm{vIoU}_i > 0.3).
\end{equation}

\noindent
Standalone tIoU and vIoU are averaged over $\mathcal{T}$ and $\mathcal{S}$, respectively, including incorrect answers.
Level-4/5 accuracy counts joint successes over all $N$ questions, and questions without the required annotations contribute zero.
Appendix~\ref{sec:ap_exp} gives the coverage-imposed bounds and scores normalized over annotated subsets. Appendix~\ref{sec:ap_protocol} compares separate and joint prediction.

\begin{table}[!t]
\centering
\caption{
Benchmark results on \textbf{\name{}} (\%). All model accuracy columns use 500 questions. Standalone temporal IoU (tIoU) and visual IoU (vIoU) are averaged over 442 and 372 annotated questions, respectively. ``Frames'' specifies sampling and frame limits. Bold and underlined values mark the best and second-best model scores. Human performance uses a randomly sampled 50-question subset. Annotation coverage and its effect on accuracy are detailed in Appendix~\ref{sec:ap_exp}.
}
\label{tab:main_results}
\renewcommand{\arraystretch}{1.0}
\setlength{\tabcolsep}{4.5pt}
\scalebox{0.95}{
\begin{tabular}{llccccccc}
\toprule
\multirow{2}{*}{\textbf{Models}} & \multirow{2}{*}{\textbf{Frames}} & \textbf{Level-1} & \textbf{Level-2} & \textbf{Level-3} & \multicolumn{2}{c}{\textbf{Level-4}} & \multicolumn{2}{c}{\textbf{Level-5}} \\
\cmidrule(lr){3-3} \cmidrule(lr){4-4} \cmidrule(lr){5-5} \cmidrule(lr){6-7} \cmidrule(lr){8-9}
& & Acc & Acc & \cellcolor{blue!8}Acc & tIoU & Acc & vIoU & \cellcolor{red!8}Acc \\
\midrule
\textit{Human Reference} & video & -- & -- & \cellcolor{blue!8}67.6 & 74.2 & 63.5 & 59.9 & \cellcolor{red!8}55.8 \\
\midrule
\multicolumn{9}{l}{\textit{Proprietary Models}} \\
\midrule
Gemini-3.7-Flash~\citep{gemini-37-flash} & video & \textbf{40.0} & \textbf{33.0} & \cellcolor{blue!8}\textbf{24.8} & \textbf{42.9} & \textbf{14.2} & 10.2 & \cellcolor{red!8}\textbf{1.8} \\
Gemini-3-Pro~\citep{gemini-3-pro}        & video & 23.6 & 20.6 & \cellcolor{blue!8}17.0 & \underline{32.0} & \underline{8.0}  &  9.6  & \cellcolor{red!8}1.0 \\
Gemini-2.5-Pro~\citep{gemini2.5}      & video & 25.2 & 21.0 & \cellcolor{blue!8}16.4 & 31.4 & 7.2  & 8.6  & \cellcolor{red!8}0.6 \\
Gemini-2.5-Flash~\citep{gemini2.5}    & video & 17.0 & 11.2 & \cellcolor{blue!8}11.4 & 27.9 & 5.4  & 4.9  & \cellcolor{red!8}0.0 \\
GPT-5.6-Sol~\citep{gpt-5.6-sol} & 1fps, 96f & \underline{32.0} & \underline{22.2} & \cellcolor{blue!8}\underline{18.8} & 23.1 & 6.4 & \underline{16.6} & \cellcolor{red!8}\underline{1.6} \\
GPT-5.2~\citep{gpt-5.2}             & 1fps, 96f    & 17.4 & 14.6 & \cellcolor{blue!8}8.0  & 21.9 & 1.8  & 12.6 & \cellcolor{red!8}0.4 \\
Seed-2.0-Pro~\citep{seed2.0}        & 1fps, 160f   & 25.4 & 18.0 & \cellcolor{blue!8}10.6 & 25.3 & 3.8  & \textbf{21.8} & \cellcolor{red!8}0.6 \\
\midrule
\multicolumn{9}{l}{\textit{Open-Source Models}} \\
\midrule
Qwen3.5-397B-A17B~\citep{qwen3.5} & 1fps, 250f  & 25.0   & 20.0   & \cellcolor{blue!8}10.6 & 3.1 & 0.0 & 8.0   & \cellcolor{red!8}0.0  \\
Qwen3-VL-235B-A22B~\citep{bai2025qwen3vl}   & 1fps, 384f  & 28.4 & 21.4 & \cellcolor{blue!8}9.6  & 19.6 & 3.4  & 3.6  & \cellcolor{red!8}0.2 \\
Qwen3-VL-8B~\citep{bai2025qwen3vl}          & 1fps, 384f  & 24.8 & 17.8 & \cellcolor{blue!8}8.2  & 10.9 & 0.6  & 2.4  & \cellcolor{red!8}0.2 \\
Qwen3-VL-4B~\citep{bai2025qwen3vl}          & 1fps, 384f  & 22.2 & 16.6 & \cellcolor{blue!8}7.8  & 13.3 & 1.4  & 1.2  & \cellcolor{red!8}0.0 \\
Qwen2.5-VL-7B~\citep{bai2025qwen25vltechnicalreport}       & 1fps, 384f  & 16.2 & 12.4 & \cellcolor{blue!8}6.0  & 1.8  & 0.0  & 1.4  & \cellcolor{red!8}0.0 \\
% Molmo2-8B~\citep{clark2026molmo2}    & 1fps, 384f  &  &  & \cellcolor{blue!8}  &   &  &  & \cellcolor{red!8} \\
InternVL3.5-8B~\citep{wang2025internvl3}       & 1fps, 96f   & 19.4 & 18.0 & \cellcolor{blue!8}9.8  & 2.9  & 0.2  & 0.8  & \cellcolor{red!8}0.0 \\
InternVL3.5-4B~\citep{wang2025internvl3}       & 1fps, 96f   & 14.8 & 13.6 & \cellcolor{blue!8}5.8  & 3.3  & 0.0  & 1.2  & \cellcolor{red!8}0.0 \\
\midrule
\multicolumn{9}{l}{\textit{Video Reasoning Specialists}} \\
\midrule
Video-R1-7B~\citep{feng2025video-r1}         & 1fps, 384f  & 17.2 & 13.4 & \cellcolor{blue!8}7.8  & 2.3  & 0.0  & 5.5  & \cellcolor{red!8}0.0 \\
VideoRFT-7B~\citep{wang2025videorft}          & 1fps, 384f  & 12.0  & 11.2  & \cellcolor{blue!8}5.8  & 3.6  & 0.0  & 4.2  & \cellcolor{red!8}0.0 \\
Open-o3-Video-4B~\citep{meng2025open}     & 1fps, 384f  & 17.6 & 12.8 & \cellcolor{blue!8}5.8  & 9.7  & 0.4  & 9.9  & \cellcolor{red!8}0.0 \\
VideoChat-R1.5-7B~\citep{yan2025videochat}    & 1fps, 384f  & 17.4 & 12.2 & \cellcolor{blue!8}8.6  & 3.3  & 0.0  & 2.9  & \cellcolor{red!8}0.0 \\
Video-o3-7B~\citep{zeng2026videoo3}    & 1fps, 384f  & 15.0 & 13.4 & \cellcolor{blue!8}4.8  & 3.5 & 0.0 & 0.2 & \cellcolor{red!8}0.0 \\
\bottomrule
\end{tabular}
}
\end{table}

\section{Evaluation on \name{}}
\label{sec:evaluation}

We conduct a comprehensive evaluation of state-of-the-art video MLLMs on \name{} under the proposed five-level hierarchical protocol.
We first present the overall experimental settings and main results across all evaluation levels, and we further perform detailed performance and error analyses from multiple perspectives to better understand capability gaps and failure patterns.

\subsection{Settings}

In Table~\ref{tab:main_results}, we evaluate 19 representative models across three categories.

\noindent
\textbf{(1) Proprietary:} Gemini-3.7-Flash~\citep{gemini-37-flash}, Gemini-3-Pro~\citep{gemini-3-pro}, Gemini-2.5~\citep{gemini2.5}, Seed-2.0-Pro~\citep{seed2.0}, GPT-5.6-Sol and GPT-5.2~\citep{gpt-5.6-sol,gpt-5.2}.

\noindent
\textbf{(2) Open-source MLLMs:} Qwen3.5~\citep{qwen3.5}, Qwen3-VL~\citep{bai2025qwen3vl}, Qwen2.5~\citep{bai2025qwen25vltechnicalreport}, and InternVL3.5~\citep{wang2025internvl3}.
%and Molmo2-8B~\citep{clark2026molmo2}.
%

\noindent
\textbf{(3) Video-specific reasoning models:} (a) textual reasoning models: Video-R1~\citep{feng2025video-r1} and VideoRFT~\citep{wang2025videorft}, (b) thinking-with-videos models: VideoChat-R1.5~\citep{yan2025videochat}, Video-o3~\citep{zeng2026videoo3}, and Open-o3-Video~\citep{meng2025open}.

We evaluate all models using the same questions and scoring rules (Section~\ref{sec:fivelevel}).
Input formats and frame budgets follow each model's supported interface and serving constraints.
Gemini models receive full videos with audio, whereas other models use uniform 1 FPS frame sampling with limits shown in Table~\ref{tab:main_results}.
In addition, since Level-5 requires predicting bounding boxes at specified timestamps, we ensure that the corresponding frames are always included in the sampled frames, with each frame mapped to its timestamp in the input metadata.
Results characterize these systems under their supported interfaces and reported budgets. Input-modality and frame-budget comparisons are provided in Tables~\ref{tab:modality} and~\ref{tab:frame_ablation}. Further details are in Appendix~\ref{sec:ap_exp}.

\subsection{Benchmark Results}

Table~\ref{tab:main_results} reports evaluation results across five levels.
Under the standard QA (Level-3) setting, Gemini-3.7-Flash achieves the highest accuracy at 24.8\%, followed by GPT-5.6-Sol at 18.8\% and Gemini-3-Pro at 17.0\%.
On a shared 50-question subset, humans reach 67.6\% QA accuracy, compared with 22.0\% and 20.0\% for Gemini-3-Pro and Gemini-2.5-Pro (Table~\ref{tab:human_study}).
Human Level-4/5 scores on that subset are 63.5\% and 55.8\%. The human study is described in Appendix~\ref{sec:ap_human}.
Among open-source models, only Qwen3.5 surpasses 10\% accuracy.
Under the strict Level-5 requirement, Gemini-3.7-Flash reaches 1.8\%, followed by GPT-5.6-Sol at 1.6\%.
Eight of the 19 models obtain non-zero scores at this level.
These results indicate that current models rarely answer these challenging questions correctly while providing accurate supporting spatio-temporal evidence.
\textbf{Answer accuracy alone does not capture the localization weaknesses observed here.}

\begin{figure}[t]
    \centering
    \includegraphics[width=0.95\linewidth]{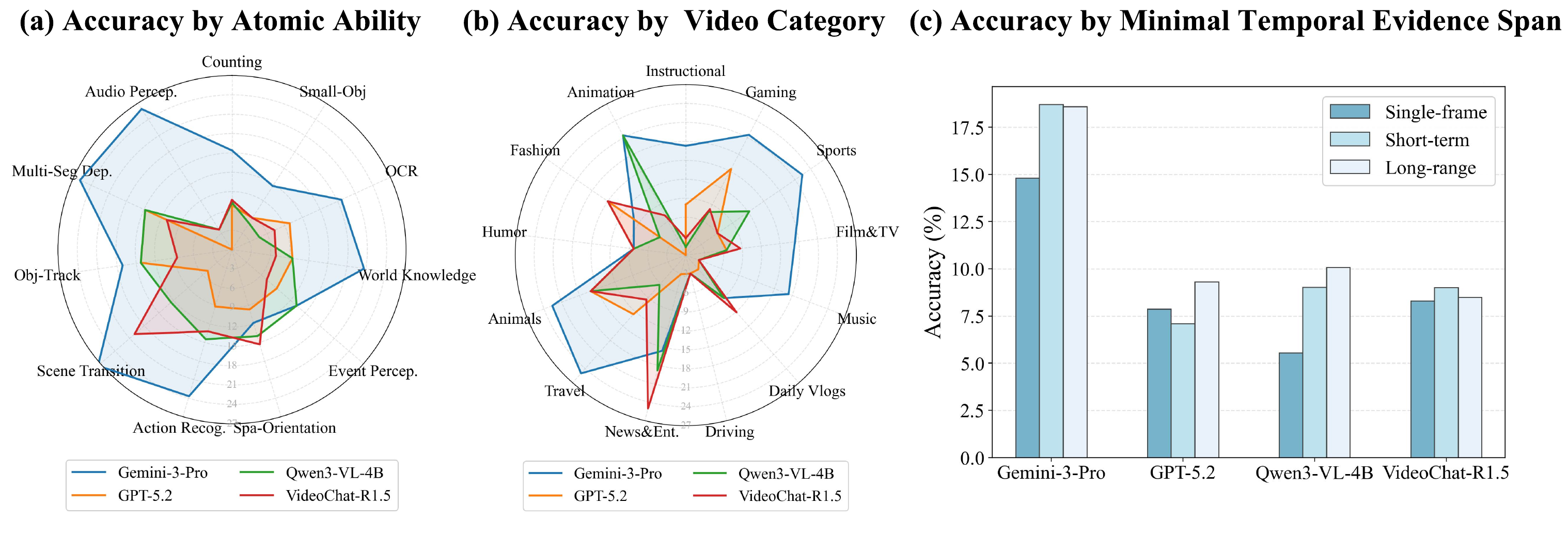}
    \caption{Performance across atomic abilities, video categories, and minimal evidence spans.}
    \label{fig:analysis}
\end{figure}

\textbf{\textit{Across evaluation levels,}} models achieve higher QA accuracy with spatio-temporal evidence hints (Level-1) than without hints (Level-3).
When spatio-temporal evidence is provided (Level-1), Gemini-3.7-Flash reaches 40.0\% and GPT-5.6-Sol reaches 32.0\%, compared with their Level-3 scores of 24.8\% and 18.8\%.
These gains show the value of evidence hints for answering.
However, even the highest-scoring model answers 60\% of questions incorrectly with evidence hints, suggesting continued difficulties in perceiving and reasoning over the indicated cues.
Removing spatial hints (Level-2) lowers accuracy relative to Level-1 across all models, showing the value of region-level information.
Notably, most current thinking-with-videos frameworks~\citep{zeng2026videoo3,ding2025videozoomer} emphasize temporal retrieval and iterative zoom-in, while systematic spatial exploration remains limited.
Our results suggest that fine-grained spatial search should also be incorporated as a core dimension alongside temporal reasoning.

Level-4 exposes a gap in temporal grounding.
Gemini-3.7-Flash leads at 14.2\%, followed by Gemini-3-Pro at 8.0\% and Gemini-2.5-Pro at 7.2\%. GPT-5.6-Sol reaches 6.4\% despite ranking second in standard QA.
Among open-source models, Qwen3-VL-235B achieves the highest Level-4 accuracy at 3.4\%. Seven of the 19 models score 0\% at this level.
In terms of temporal grounding, Gemini-3.7-Flash reaches 42.9\% tIoU, followed by Gemini-3-Pro at 32.0\% and Gemini-2.5-Pro at 31.4\%. Most models remain below 20\%.
These results indicate limited capability to identify correct temporal evidence in complex scenes.
At Level-5, models are required to perform spatial grounding on key frames.
Seed-2.0-Pro achieves the highest vIoU of 21.8\%, followed by GPT-5.6-Sol at 16.6\% and GPT-5.2 at 12.6\%.
Gemini-3.7-Flash reaches 10.2\% vIoU despite leading Level-5 accuracy, showing that QA and standalone localization scores capture different strengths.
On the 50-question human subset, the reference tIoU and vIoU are 74.2\% and 59.9\%, respectively.
As shown in Figure~\ref{fig:cases}~(2), the Gemini model produces the correct answer while providing inaccurate bounding boxes.

\textbf{\textit{Across model categories,}} most proprietary systems outperform open-source models under the evaluated input settings.
Gemini-3.7-Flash achieves the highest accuracy at all five levels and the highest temporal IoU, while Seed-2.0-Pro achieves the highest spatial IoU. GPT-5.6-Sol ranks second at Levels 1--3 and 5.
Among open-source models, Qwen3-VL-235B leads in tIoU (19.6\%), Level-4 accuracy (3.4\%), Level-1 (28.4\%), and Level-2 (21.4\%), while Qwen3.5 leads standard QA (10.6\%).
These differences motivate reporting answering and localization scores separately.

For video-specific reasoning models, none achieves non-zero performance at Level-5, and only Open-o3-Video obtains non-zero Level-4 accuracy.
In terms of grounding capability, it also achieves higher temporal and spatial localization scores than other reasoning models, with 9.7\% tIoU and 9.9\% vIoU.
For agentic thinking-with-videos models such as VideoChat-R1.5 and Video-o3, although they adopt multi-round reasoning and iterative zoom-in strategies, their performance remains 0 at both Level-4 and Level-5, showing no clear advantage on the leaderboard.
For these tested systems, \textbf{iterative evidence seeking has yet to yield high grounding accuracy on this benchmark.}

\begin{table*}[!t]
\centering
\begin{minipage}[t]{0.44\textwidth}
\centering
\caption{\small Effect of the thinking-with-videos paradigm. Here, L1–L4 denote accuracy under the first four levels on VideoChat-R1.5.}
\label{tab:agentic}
\resizebox{\linewidth}{!}{
\begin{tabular}{lcccc}
\toprule
\textbf{VideoChat-R1.5} & \textbf{L1} & \textbf{L2} & \textbf{L3}  & \textbf{L4}\\
\midrule
Three Rounds & 17.4 & 12.2 & 8.6 & 0.0\\
Single Round   & 16.0 & 11.2 & 6.8 & 0.0\\
\bottomrule
\end{tabular}
}
\end{minipage}
\hspace{0.01\textwidth}
\begin{minipage}[t]{0.49\textwidth}
\centering
\small
\caption{\small Comparison of visual input: original video, temporal segments only (zoom), and temporal segments with spatial crops (zoom\&crop).}
\label{tab:evidence}
\scalebox{0.77}{
\begin{tabular}{lccc}
\toprule
\textbf{Model} & \textbf{Ori-Video} & \textbf{Zoom} & \textbf{Zoom\&Crop} \\
\midrule
Gemini-3-Pro   & 17.0 & 21.6 & 41.2 \\
Gemini-2.5-Pro & 16.4 & 21.0 & 42.0 \\
Qwen3-VL-4B    & 7.8  & 11.6 & 18.0 \\
\bottomrule
\end{tabular}
}
\end{minipage}
\end{table*}

\subsection{Findings and Discussion}

\begin{figure}[!t]
\centering
\includegraphics[width=1.0\linewidth]{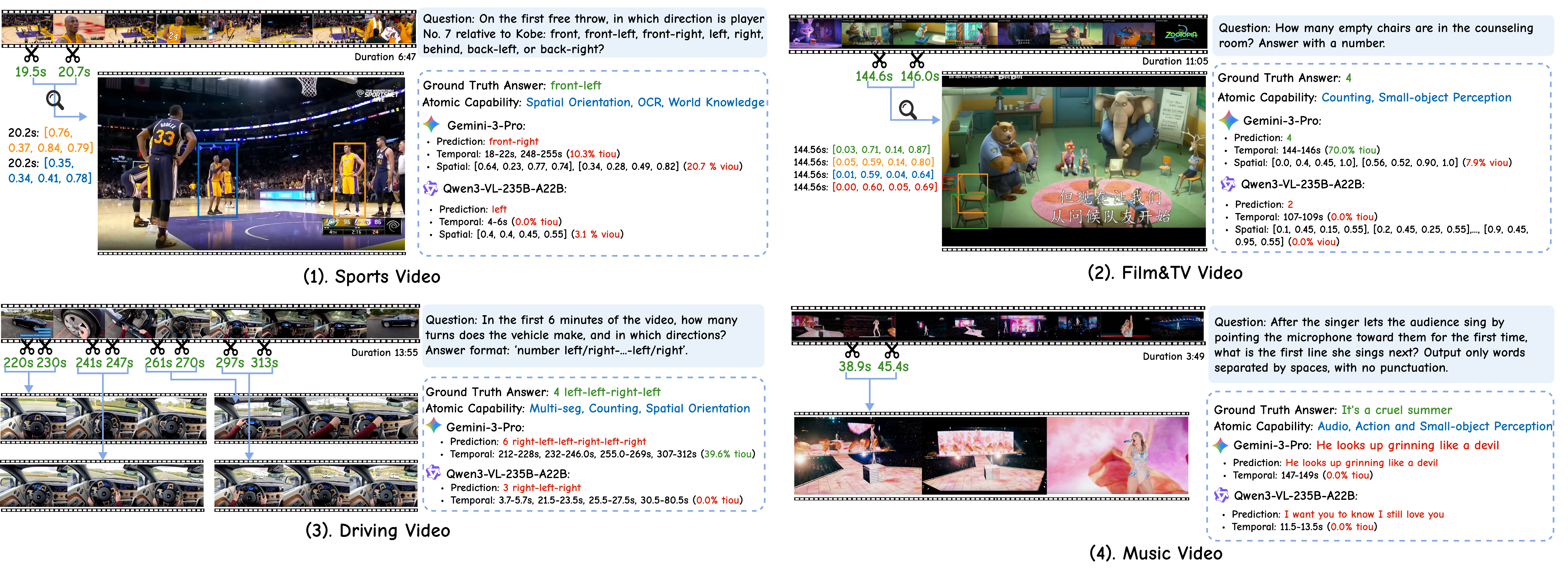}
\caption{
\small
Examples from \name{} with annotated spatio-temporal evidence and model predictions.
(1) Models localize redundant or incorrect evidence and misinterpret relative spatial relationships.
(2) Correct answers may still lack precise spatial grounding for all relevant objects.
(3) Models fail to integrate evidence across multiple temporal segments and correctly determine turning directions.
(4) Models struggle to combine auditory cues with visual evidence for reasoning.
}
\label{fig:cases}
\end{figure}

\noindent
\textbf{Q1: Which atomic abilities remain challenging for current video MLLMs?}
We analyze atomic abilities for the models shown in Figure~\ref{fig:analysis}~(left).
There are two key observations.
\textbf{I. Small-object perception and spatial orientation discrimination remain particularly challenging.}
For example, Gemini-3-Pro achieves 11.7\% on small-object perception and 11.8\% on spatial orientation, both below its overall QA accuracy.
From Figure~\ref{fig:analysis}~(middle), the Driving category also exhibits low accuracy.
The driving questions test subtle targets and relative positions in cluttered scenes. Their low accuracy is consistent with the spatial difficulties illustrated in the case studies.
As shown in Figure~\ref{fig:cases}~(2), models frequently fail to localize the correct targets, and in Figure~\ref{fig:cases} (1) and (3), models also misinterpret spatial direction in realistic environments.
\textbf{II. Counting is consistently difficult across models.}
Counting requires accurate localization of evidence and cross-frame aggregation.
It involves both intra-frame object counting and inter-frame event counting.
Models need to distinguish repeated appearances and handle occlusion.
Empirically, most models shown achieve accuracies below 8\% for counting.
This reflects limited spatial precision and weak integration of multi-segment evidence.
As shown in Figure~\ref{fig:cases}~(2)(3), models often fail to fully identify small objects or short-lived actions.

\noindent
\textbf{Q2: Are questions answerable from a single frame inherently easier?}
Some video questions can be answered from a single frame, requiring minimal temporal coverage.
Without temporal reasoning, can models handle such problems more easily?
To investigate this, we categorize questions by minimal temporal evidence span.
As shown in Figure~\ref{fig:analysis}~(right), for the models included in this analysis, single-frame questions are not easier than short-term or long-range ones.
As illustrated in Figure~\ref{fig:cases}~(2), such questions in our benchmark often resemble needle-in-a-haystack scenarios.
Although a single frame may suffice to answer the question, the relevant visual cue may be visible for only a few seconds.
Uniform sampling may miss such fleeting frames, limiting the evidence available for answering.
\textbf{Therefore, with a limited visual token budget and constrained frame sampling, precise temporal localization of short-lived yet critical frames remains highly challenging.}

\noindent
\textbf{Q3: Does the agentic thinking-with-videos paradigm bring gains?}
We conduct an ablation study using VideoChat-R1.5 as a representative thinking-with-videos model.
This paradigm alleviates the limitations of uniform frame sampling by iteratively refining temporal focus.
We compare three-round zoom-in reasoning with single-round direct answering.
Under this paradigm, in round $(n-1)$ the model outputs its reasoning trace, predicted answer, and candidate temporal intervals, and in round $n$ it performs dense sampling within the predicted interval and sparse sampling over the remaining video.
As shown in Table~\ref{tab:agentic}, accuracy improves by 1.4, 1.0, and 1.8 percentage points at Level-1, Level-2, and Level-3, respectively.
These QA gains do not translate into improved Level-4/5 accuracy.
\textbf{Fine-grained grounding remains difficult for VideoChat-R1.5 under this thinking-with-videos setting.}

\noindent
\textbf{Q4: What is the gap between full-video and visual-evidence-only input?}
Beyond the standard Level-1 and Level-2 settings, we further compare different visual input configurations while controlling total visual tokens.
As shown in Table~\ref{tab:evidence}, we evaluate three conditions. The first uses the original full video without hints (Level-3). The second only uses annotated temporal segments with zoom-in sampling. The third uses annotated segments together with cropped key spatial regions.
For Gemini models, zoom and crop substantially improve performance. \textbf{The accuracies of Gemini-3-Pro and Gemini-2.5-Pro increase from 17.0\% and 16.4\% to above 40\% under the zoom\&crop setting.}
These gains show the value of direct access to fine-grained evidence, motivating better temporal search and spatial inspection.

\begin{table}[!t]
\centering
\caption{Effect of input modality on Gemini-3-Pro. Small-Obj and Audio denote Level-3 performance on small-object and audio perception, respectively.}
\label{tab:modality}
\scalebox{0.95}{
\begin{tabular}{lccc}
\toprule
\textbf{Input} & \textbf{Level-3} & \textbf{Small-Obj} & \textbf{Audio} \\
\midrule
No vis.\& aud. & 6.8  & 5.4  & 0.0 \\
Audio-only     & 8.4  & 4.9  & 7.4 \\
Frames         & 17.0 & 15.6 & 7.4 \\
Video          & 17.0 & 11.7 & 25.9 \\
\bottomrule
\end{tabular}}
\end{table}

\noindent
\textbf{Q5: How does input modality affect performance?}
\label{sec:input_modality}
We compare Gemini-3-Pro with no visual or audio input, audio-only input, frames sampled at 1 FPS with a 384-frame cap, and full video with audio (Table~\ref{tab:modality}).
Frames and full video both yield 17.0\% overall QA accuracy, but their strengths differ across abilities. Full video raises audio accuracy from 7.4\% to 25.9\%, while frames yield higher small-object accuracy (15.6\% versus 11.7\%).
Without visual or audio input, the model still answers 6.8\% of questions correctly, suggesting that question priors or world knowledge can contribute to answer-only scores.
These results reinforce the need to verify supporting evidence, not just answers.

\noindent
\textbf{Additional experimental results, visualizations, and details.}
Low grounding accuracy persists under relaxed thresholds for Gemini-3-Pro (Table~\ref{tab:threshold_sensitivity}) and under joint prediction for the three models tested on a shared subset (Table~\ref{tab:joint_protocol}).
Appendices~\ref{sec:ap_analysis}, \ref{sec:ap_human}, \ref{sec:ap_reliability}, and~\ref{sec:ap_vis} provide further ablations, the human study, repeated-run results, and error cases.
% Keep all experiment floats inside the evaluation section instead of drifting past the bibliography.
\FloatBarrier

\section{Conclusion}
\label{sec:conclusion}

We present \textbf{\name{}}, a challenging hierarchical benchmark for evidence-grounded long-video question answering.
Its five-level diagnostic protocol evaluates answering under different evidence hints and combines answer correctness with temporal and spatial localization scores.
The best model reaches 24.8\% in standard QA and 1.8\% with correct answers and accurate spatio-temporal evidence.
Our evaluation yields three takeaways: (1) correct answers do not ensure accurate evidence localization; (2) direct access to fine-grained visual evidence substantially improves QA; and (3) the tested thinking-with-videos approaches still struggle with precise grounding.
These findings motivate methods that combine temporal search, spatial inspection, and explicit evidence verification.
Limitations in coverage and evidence granularity, together with future benchmark extensions, are discussed in Appendix~\ref{sec:ap_limitations}.
We hope \name{} can help advance video MLLMs that answer challenging long-video questions with precise, reliable, and verifiable evidence.

\noindent
\textbf{Acknowledgement.} 
This work is supported by the National Key Research and Development Program of China (No. 2023YFC3807600).

\bibliography{main}
\bibliographystyle{plainnat}

\clearpage
\appendix

\section{Appendix for VideoZeroBench}

\noindent
\textbf{Overview.}
This appendix provides additional implementation details, annotation descriptions, and extended analyses to complement the main paper.
\begin{itemize}
  \setlength{\itemsep}{0.2em}
  \setlength{\parskip}{0pt}
  \setlength{\parsep}{0pt}
  \item Section~\ref{sec:ap_exp} presents additional experimental details.
  \item Section~\ref{sec:ap_analysis} provides further analyses of frame sampling strategies, test-time scaling, the effect of video length, and the sensitivity of results to tIoU and vIoU thresholds.
  \item Section~\ref{sec:ap_case_comparison} provides a detailed comparison with recent video spatio-temporal grounding benchmarks.
  \item Section~\ref{sec:ap_data} describes additional annotation details.
  \item Section~\ref{sec:ap_human} reports human performance at Levels 3--5 on a subset of \name{}.
  \item Section~\ref{sec:ap_protocol} explains the rationale behind our evaluation protocol.
  \item Section~\ref{sec:ap_reliability} examines the robustness of our findings.
  \item Section~\ref{sec:ap_vis} presents additional qualitative visualizations and error cases to illustrate typical failure patterns of current video MLLMs.
  \item Section~\ref{sec:ap_related_works} discusses related work.
  \item Section~\ref{sec:ap_limitations} discusses limitations and future work.
\end{itemize}

\newpage

\subsection{Additional Experimental Details}
\label{sec:ap_exp}

This section provides additional experimental details for reproducibility.

\noindent
\textbf{Questions and annotation coverage.}
The benchmark contains 500 distinct question--answer pairs.
Table~\ref{tab:annotation_coverage} reports the availability of sparse manual evidence annotations.
There are 442 questions with temporal intervals and 372 with key-frame boxes, of which 320 have both.
The reported 2,314 task instances count the three QA settings over the shared question set, plus 442 temporal and 372 spatial annotation targets: $3\times500+442+372=2{,}314$.
This instance count distinguishes repeated tasks from distinct questions.

\begin{table}[!ht]
\centering
\small
\caption{Sparse manual annotation coverage in \name{}.}
\label{tab:annotation_coverage}
\begin{tabular}{lrrr}
\toprule
 & Spatial boxes & No spatial boxes & Total \\
\midrule
Temporal intervals & 320 & 122 & 442 \\
No temporal intervals & 52 & 6 & 58 \\
\midrule
Total & 372 & 128 & 500 \\
\bottomrule
\end{tabular}
\end{table}

\noindent
\textbf{Video Input and Frame Sampling.}
For the Gemini series, the raw video is base64-encoded and sent directly to the API as input. Due to API size limits, videos are compressed to less than 100MB before encoding.
All other models require sampled frames as input. Frames are uniformly sampled at 1 FPS with a predefined maximum number of frames.
As shown in Table~\ref{tab:main_results}, most models use a limit of 384 frames.
Some models use smaller limits due to context length or memory constraints, including Seed-2.0-Pro~\citep{seed2.0} (160 frames), Qwen3.5-397B~\citep{qwen3.5} (250 frames), and GPT-5.6-Sol~\citep{gpt-5.6-sol}, GPT-5.2~\citep{gpt-5.2}, and InternVL3.5~\citep{wang2025internvl3} (96 frames each).
Each sampled frame is resized while preserving the original aspect ratio.
For Level-5, the visual input explicitly includes every designated key frame and its timestamp, including when raw video is supplied.
These settings aim to make use of each model's supported input capabilities. We use native video with audio for Gemini and sampled frames for the other models.
Section~\ref{sec:ap_analysis} (Table~\ref{tab:frame_ablation}) reports frame-rate and frame-budget ablations. Table~\ref{tab:modality} in the main paper reports the input-modality ablation.

\noindent
\textbf{Inference Setup.}
Proprietary models and Qwen3.5-397B~\citep{qwen3.5} are evaluated through API calls. All other open-source models are evaluated using the vLLM inference framework.
The decoding temperature is set to 0 for the main evaluation. Repeated-run results at temperature 0.7 are reported in Section~\ref{sec:ap_reliability}.
Models with native chain-of-thought generation keep their default reasoning style during evaluation, while other models are prompted to output only the final answer.

\noindent
\textbf{Thinking-with-videos Paradigm Configurations.}
For VideoChat-R1.5~\citep{yan2025videochat}, frames are sampled more densely within predicted key intervals than in other parts of the video. The density ratio is controlled by the parameter \texttt{key\_ratio}. We follow the original setting and use $\texttt{key\_ratio}=1.5$. Multi-round reasoning is fixed at three rounds.
For Video-o3~\citep{zeng2026videoo3}, we follow the original paper and use dynamic sampling with adaptive pixel budgets. The reasoning process runs for at most eight rounds. If the maximum round limit is reached, the model is forced to output the final answer.

\begin{figure}[t]
    \centering
    \includegraphics[width=0.97\linewidth]{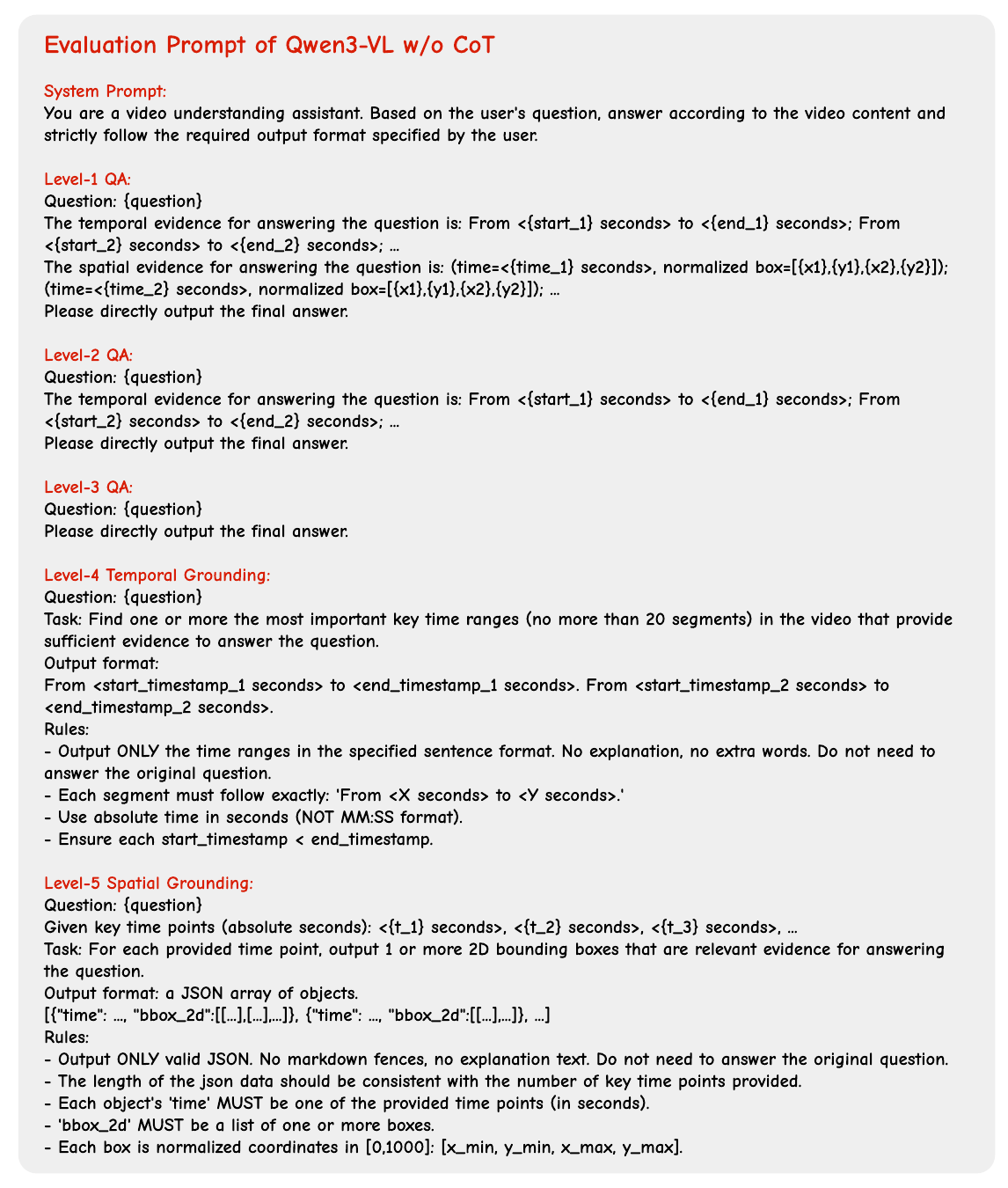}
    \caption{Example prompt templates used for Qwen3-VL models across different evaluation levels.}
    \label{fig:prompt_templates}
\end{figure}

\noindent
\textbf{Prompt Templates Across Five Levels.}
Prompt templates are adapted to match the timestamp and coordinate formats required by different models.
For example, for Qwen3-VL models, bounding boxes are represented using normalized coordinates in the range 0--1000.
For models based on Qwen2.5-VL, bounding boxes are represented using absolute pixel coordinates.
The output parser is implemented accordingly to correctly extract temporal intervals and spatial boxes.
Figure~\ref{fig:prompt_templates} illustrates example prompt templates used for Qwen3-VL models (without chain-of-thought) across different evaluation levels.

\noindent
\textbf{Score Computation.}
All five accuracy scores use the full question pool as a common denominator ($N=500$).
This choice expresses Level-4/5 joint success and Level-1--3 answering success on the same benchmark-wide scale, making their gaps easier to compare without changing the reference population.
For Level-4/5 accuracy, questions lacking the required annotations contribute no verified joint success. Standalone tIoU and vIoU are instead averaged over the 442 temporally and 372 spatially annotated questions, respectively, regardless of answer correctness.
The annotation coverage therefore bounds Level-4 accuracy by $442/500=88.4\%$ and Level-5 accuracy by $320/500=64.0\%$.
Cross-level accuracy gaps reflect both annotation coverage and prediction failures.

Re-normalizing the best model's existing success counts over the relevant annotated subsets gives $71/442=16.1\%$ for Level-4 and $9/320=2.8\%$ for Level-5, compared with the reported 14.2\% and 1.8\% on the full question pool.
Thus, low joint success persists on the annotated subsets. This change of denominator also preserves model rankings within each level.

We use an IoU threshold of 0.3 for both temporal and spatial grounding.
This threshold is chosen to account for the inherent ambiguity of video evidence annotation:
supporting temporal intervals may have soft boundaries, and spatial boxes may vary slightly across annotators,
especially for small objects, motion blur, and cluttered scenes.
At the same time, the threshold still requires the predicted evidence to substantially overlap with the annotated supporting region,
preventing models from receiving credit for coarse or irrelevant localization.
If a model fails to output a valid temporal interval or spatial box, the accuracy score for this query is set to 0 in Level-4 or Level-5.
Temporal IoU measures precision at the annotated temporal granularity: a substantially wider prediction can cover the event yet fall below the threshold.
Accordingly, low tIoU can reflect imprecise boundaries as well as an incorrect event location. Table~\ref{tab:threshold_sensitivity} examines threshold sensitivity.

\subsection{Additional Analysis}
\label{sec:ap_analysis}

This section examines frame sampling, test-time scaling, video length, and grounding thresholds. The input-modality experiment is reported in the main paper (Table~\ref{tab:modality}).

\noindent
\textbf{Effect of Frame Sampling Strategy.}
We analyze frame sampling for Qwen3-VL-4B and Qwen3-VL-235B by varying the maximum number of input frames.
Table~\ref{tab:frame_ablation} reports the Level-3 accuracy under different frame budgets.
The first four settings use 1 FPS with different frame caps, while the last setting uses 2 FPS with a maximum of 768 frames.
The results show that increasing the frame budget does not consistently improve performance.
Both models reach their highest tested accuracy at 96 frames, with no consistent gains at larger budgets.
This motivates more selective temporal sampling for these models rather than relying only on more frames.
Future work should instead focus on identifying informative temporal segments through temporal grounding, which increases sampling density in key segments while reducing redundant frames.

\noindent
\textbf{Effect of Test-Time Scaling.}
We investigate whether allocating more inference compute at test time improves long-video understanding for Qwen3-VL-4B~\citep{bai2025qwen3vl} under the standard Level-3 setting.
We evaluate two distinct paradigms.
The first is \textit{parallel scaling}, where we sample $K=5$ independent responses per question at a higher temperature ($\tau=0.7$) to determine whether the model can generate the correct answer among multiple attempts.
The second is a two-round \textit{sequential scaling} approach (localize-then-answer): In the first round, the model predicts key temporal evidence windows.
In the second round, the video is dynamically resampled with higher density within these predicted windows to generate the final answer.

As shown in Table~\ref{tab:test_time_scaling}, parallel scaling notably increases the upper bound of correctness, with Pass@5 (Any) reaching 15.0\% compared to the 7.8\% greedy decoding baseline. However, the extremely low Pass@5 (All) score of 1.8\% reveals high uncertainty in the model's reasoning paths.
Conversely, the sequential reasoning paradigm degrades performance to 7.0\%. This decline aligns with our previous findings regarding the temporal grounding bottleneck:
Because the model's intrinsic localization ability is weak, the first round frequently predicts incorrect evidence windows.
The second round may concentrate its sampling budget on irrelevant frames, filtering out genuine visual cues.

\begin{table}[!t]
\centering
\caption{\textbf{Effect of test-time scaling strategies on Qwen3-VL-4B under the Level-3 setting.} The baseline utilizes single-round greedy decoding. Pass@5 (Any) indicates that at least one of the 5 generated answers is correct, while Pass@5 (All) requires all 5 answers to be correct.}
\label{tab:test_time_scaling}
\begin{tabular}{lc}
\toprule
Method & Accuracy (\%) \\
\midrule
Baseline (Greedy Decoding) & 7.8 \\
\midrule
\multicolumn{2}{l}{\textit{Parallel Scaling (Pass@5, $\tau=0.7$)}} \\
Pass@5 (Any) & 15.0 \\
Pass@5 (All) & 1.8 \\
\midrule
\multicolumn{2}{l}{\textit{Sequential Scaling}} \\
Localize-then-Answer & 7.0 \\
\bottomrule
\end{tabular}
\end{table}

\begin{figure}[t]
\centering
\includegraphics[width=0.85\linewidth]{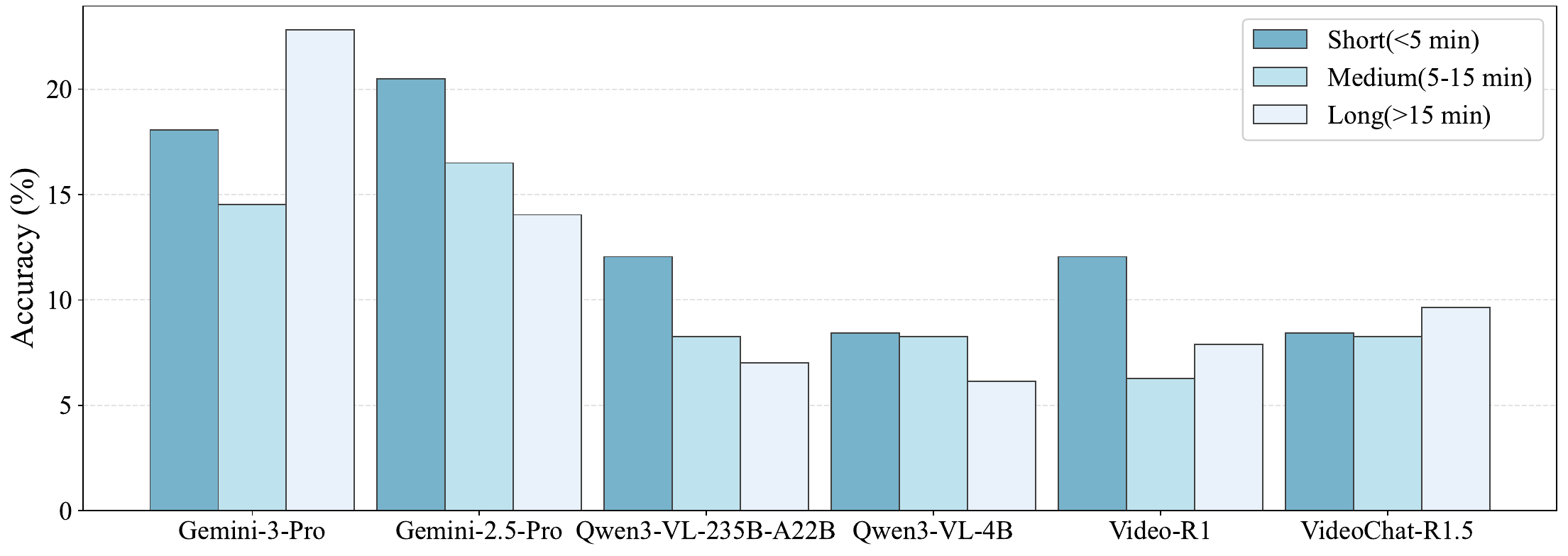}
\caption{Level-3 accuracy grouped by video duration. Videos are divided into short ($<$5 minutes), medium (5--15 minutes), and long ($>$15 minutes).}
\label{fig:duration_analysis}
\end{figure}

\noindent
\textbf{Effect of Video Length.}
We further analyze model performance under different video durations.
Videos are grouped into three categories based on their length: short ($<$5 minutes), medium (5--15 minutes), and long ($>$15 minutes).
Figure~\ref{fig:duration_analysis} shows the Level-3 accuracy for each group.
For most models shown, accuracy decreases as the video duration increases.
This trend suggests that long videos introduce additional challenges for current video MLLMs, including a larger search space, more distractors, and more complex reasoning logic.
There are also some exceptions.
Gemini-3-Pro~\citep{gemini-3-pro} and video reasoning models such as Video-R1~\citep{feng2025video-r1} and VideoChat-R1.5~\citep{yan2025videochat} demonstrate stronger robustness on long videos.
This behavior indicates that advanced reasoning capabilities may help mitigate the difficulty of long-video comprehension.

\noindent
\textbf{Threshold Sensitivity of tIoU and vIoU.}
We use tIoU $>$ 0.3 and vIoU $>$ 0.3 as the default setting because this threshold provides a moderate balance between evidence specificity and localization tolerance, especially for short temporal evidence and small spatial regions, as mentioned in Section~\ref{sec:ap_exp}.
As shown in Table~\ref{tab:threshold_sensitivity}, relaxing the thresholds increases Level-5 accuracy, while stricter thresholds further reduce the score.
However, Gemini-3-Pro still achieves only 3.4\% accuracy under the most relaxed setting, showing that its grounding weakness persists across the tested thresholds.

\begin{table}[t]
\centering
\caption{
\textbf{Threshold sensitivity analysis on Gemini-3-Pro.}
We report Level-5 accuracy under different combinations of temporal IoU and visual IoU thresholds.
The default setting used in the main paper is tIoU $>$ 0.3 and vIoU $>$ 0.3.
}
\label{tab:threshold_sensitivity}
\scalebox{0.98}{
\begin{tabular}{lccc}
\toprule
\multirow{2}{*}{tIoU Threshold}
& \multicolumn{3}{c}{vIoU Threshold} \\
\cmidrule(lr){2-4}
& $>$ 0.1 & $>$ 0.3 & $>$ 0.5 \\
\midrule
$>$ 0.1 & 3.4 & 1.8 & 0.6 \\
$>$ 0.3 & 2.0 & \textbf{1.0} & 0.2 \\
$>$ 0.5 & 1.2 & 0.6 & 0.2 \\
\bottomrule
\end{tabular}
}
\end{table}

\subsection{Detailed Comparisons with Related Spatio-temporal Grounding Benchmarks}
\label{sec:ap_case_comparison}

\begin{table}[t]
\centering
\small
\setlength{\tabcolsep}{3.2pt}
\renewcommand{\arraystretch}{1.18}
\caption{Comparison with recent spatio-temporal grounding benchmarks. The \name{} query count denotes distinct questions, and its five-level protocol yields 2,314 task instances.}
\label{tab:ap_benchmark_comparison}
\resizebox{\linewidth}{!}{
\begin{tabular}{p{2.8cm} p{2.8cm} p{2.6cm} p{1.45cm} p{1.2cm} p{1.6cm} p{6.0cm}}
\toprule
\textbf{Benchmark} & \textbf{Primary goal} & \textbf{Video/Data source} & \textbf{\#Videos} & \textbf{Dur.} & \textbf{\#Queries} & \textbf{Key distinctions / features} \\
\midrule
V-STaR~\citep{VSTaR} & Reverse spatial and temporal reasoning over ``what--when--where'' chains & VidSTG, TVQA+, GOT-10K, and inserted YouTube videos & 743 & 110.23s & 2.1K & Focuses on grounding over relatively salient object-event relations; lacks evaluation on difficult QA in long, complex videos and does not provide hierarchical analysis. \\
\midrule
ToG-Bench~\citep{xu2025tog} & Task-oriented STVG for embodied agents & ScanNet egocentric indoor trajectories & 100 & 87.88s & 2.7K & Focuses on task-relevant indoor object localization rather than difficult open-ended video QA; the scenes are mostly egocentric indoor environments with limited duration. \\
\midrule
Dr.V~\citep{luo2025dr} & Hallucination diagnosis across perceptive, temporal, and cognitive levels & 15 existing public video QA datasets & 4,974 & 0--600s & 10K & Targets hallucination diagnosis with many MCQ/binary/caption-style samples; most videos are relatively short and transformed from existing datasets, and the overall QA difficulty is lower. \\
\midrule
Know-Show~\citep{sugandhika2025know} & Action-conditioned grounded reasoning over persons, objects, hands, and interactions & Charades, Action Genome, Ego4D test splits & 675 & 47.97s & 2.5K & Evaluates action-conditioned entity grounding, but videos are much shorter and scenarios are tied to established action datasets with constrained action-role bindings. \\
\midrule
\textbf{\name{} (ours)} & Evidence-grounded long-video QA with strict answer and evidence verification & YouTube and Bilibili videos across 13 domains & 138 & 667.1s & 500 & Long and complex Internet videos; subtle evidence such as fleeting frames, small objects, OCR cues, and multi-segment dependencies; five-level diagnostic hierarchy; human annotations with two-round cross-verification. \\
\bottomrule
\end{tabular}}
\end{table}

\begin{figure}[t]
\centering
\includegraphics[width=1.0\textwidth]{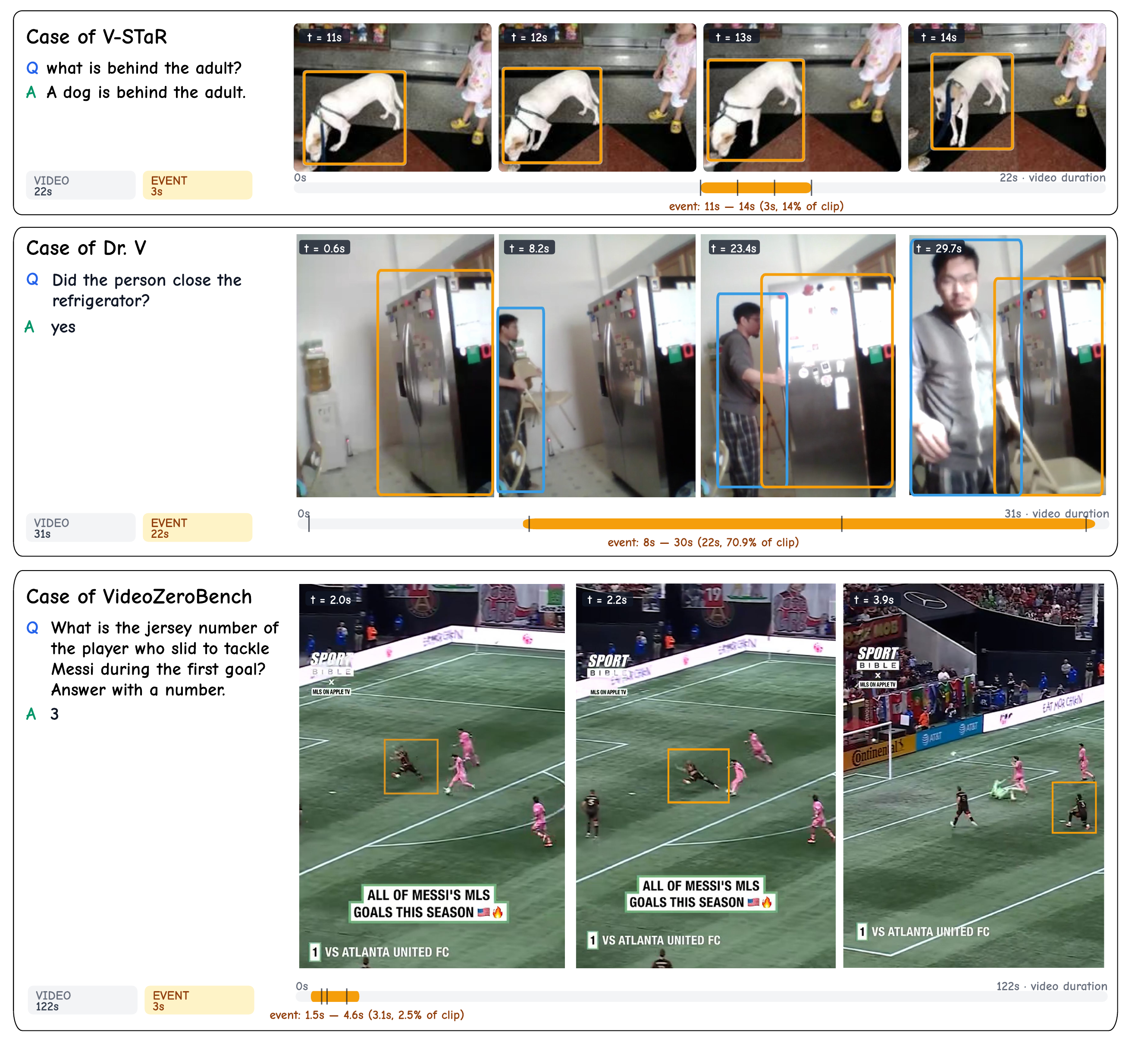}
\caption{
Qualitative comparison with representative spatio-temporal grounding benchmarks.
We randomly select examples from V-STaR, Dr.V, and \name{} for visualization.
In V-STaR, many questions focus on relatively salient targets and explicit object relations, such as a dog behind a person.
In Dr.V, many cases also examine simple actions involving obvious objects, such as whether a person closes a refrigerator.
Moreover, these videos are relatively shorter and the supporting evidence is temporally denser.
By contrast, \name{} often requires identifying events involving small objects within fleeting moments, such as a jersey number during a short tackle event, and further combines such visual evidence with higher-level semantic reasoning. The example illustrates that \name{} presents higher video complexity, harder questions, and sparser evidence.
}
\label{fig:ap_example_comparison}
\end{figure}

Recent video understanding benchmarks with spatio-temporal grounding provide valuable evidence annotations, but their evaluation goals and difficulty profiles differ from those of \name{}, as summarized in Table~\ref{tab:ap_benchmark_comparison}.
V-STaR~\citep{VSTaR} studies chain-style ``what--when--where'' reasoning, ToG-Bench~\citep{xu2025tog} focuses on task-level object localization in egocentric indoor scenes, Dr.V~\citep{luo2025dr} uses spatio-temporal evidence for hallucination diagnosis, and Know-Show~\citep{sugandhika2025know} evaluates action-conditioned entity grounding.
As illustrated in Figure~\ref{fig:ap_example_comparison}, examples from V-STaR and Dr.V often involve salient targets or simple actions over relatively short videos, making the required evidence generally more accessible once the relevant frames are observed.
In contrast, \name{} targets hard long-video QA and explicitly separates three abilities that are often mixed together in prior settings: retrieving the correct temporal evidence, localizing the relevant spatial region, and reasoning from the retrieved evidence.
This is achieved through a five-level hierarchy that starts from answer correctness under different evidence hints and gradually tightens the requirement to temporal and spatial grounding, so the gaps across levels reveal where models fail.
\name{} also introduces greater video-level complexity: its newly collected Internet videos average 667.1 seconds and contain more scene changes, distractors, and long-range dependencies than most existing spatio-temporal grounding benchmarks.
Separately, the required evidence is often intrinsically harder to perceive. Instead of relying mainly on salient large objects or obvious actions~\citep{VSTaR}, many questions depend on fleeting frames, small objects, subtle spatial relations, OCR cues, object tracks, or evidence distributed across multiple segments.
Finally, all final questions, answers, temporal intervals, and spatial boxes are manually created or verified, improving label quality and reducing the risk of contamination from older public datasets or machine-generated labels.
As a result, \name{} exposes a much sharper capability gap: even state-of-the-art models perform poorly under standard QA and nearly fail when correct answers must be supported by accurate spatio-temporal evidence.

\subsection{Additional Annotation Details}
\label{sec:ap_data}

This section covers annotation costs, filtering statistics, and the reference-generation interface.

\noindent
\textbf{Annotation Effort and Filtering.}
Annotating each QA pair and its spatio-temporal evidence takes 5--10 minutes.
Each question undergoes two verification rounds, taking 3--5 minutes per round.
We remove ambiguous or unanswerable questions and those with non-unique answers or annotation errors.
Of 608 candidate questions, 108 (17.8\%) are discarded, leaving 500.
The total annotation effort is approximately 200 person-hours.

\begin{figure}[t]
    \centering
    \includegraphics[width=0.96\linewidth,height=0.80\textheight,keepaspectratio]{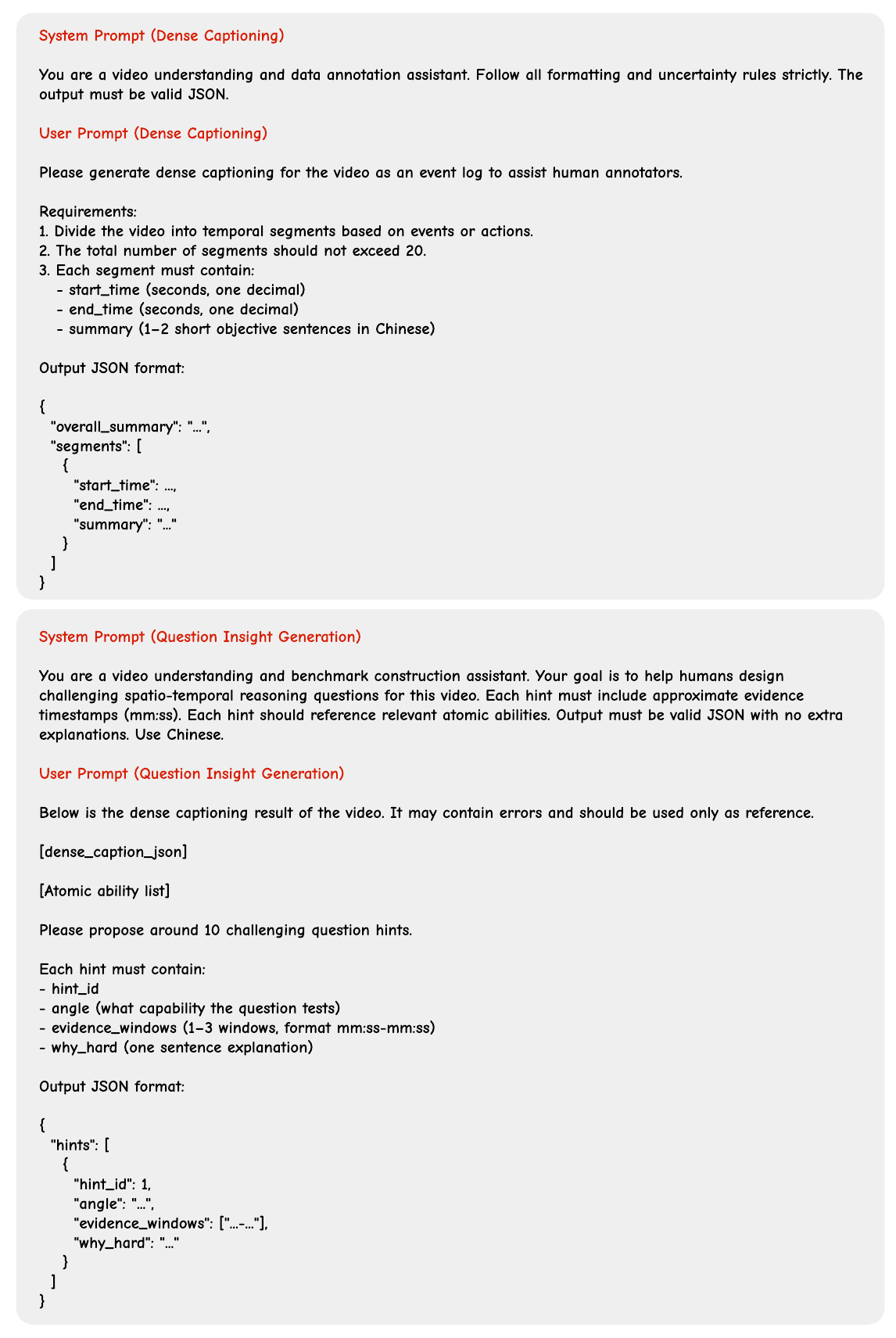}
    \caption{Prompt used to generate dense captions and question insights for annotation reference.}
    \label{fig:annotation_prompt}
\end{figure}

\noindent
\textbf{LLM-Assisted Reference Information Generation.}
Before manual annotation, we use Gemini to generate two types of auxiliary information for each video: dense captions and question insights.
These outputs help annotators quickly understand the video content and explore potential question directions.
They serve only as reference information, and all final questions and answers are designed and verified by human annotators.
Figure~\ref{fig:annotation_prompt} shows the prompt for these auxiliary descriptions.

\begin{figure}[t]
    \centering
    \includegraphics[width=0.9\linewidth]{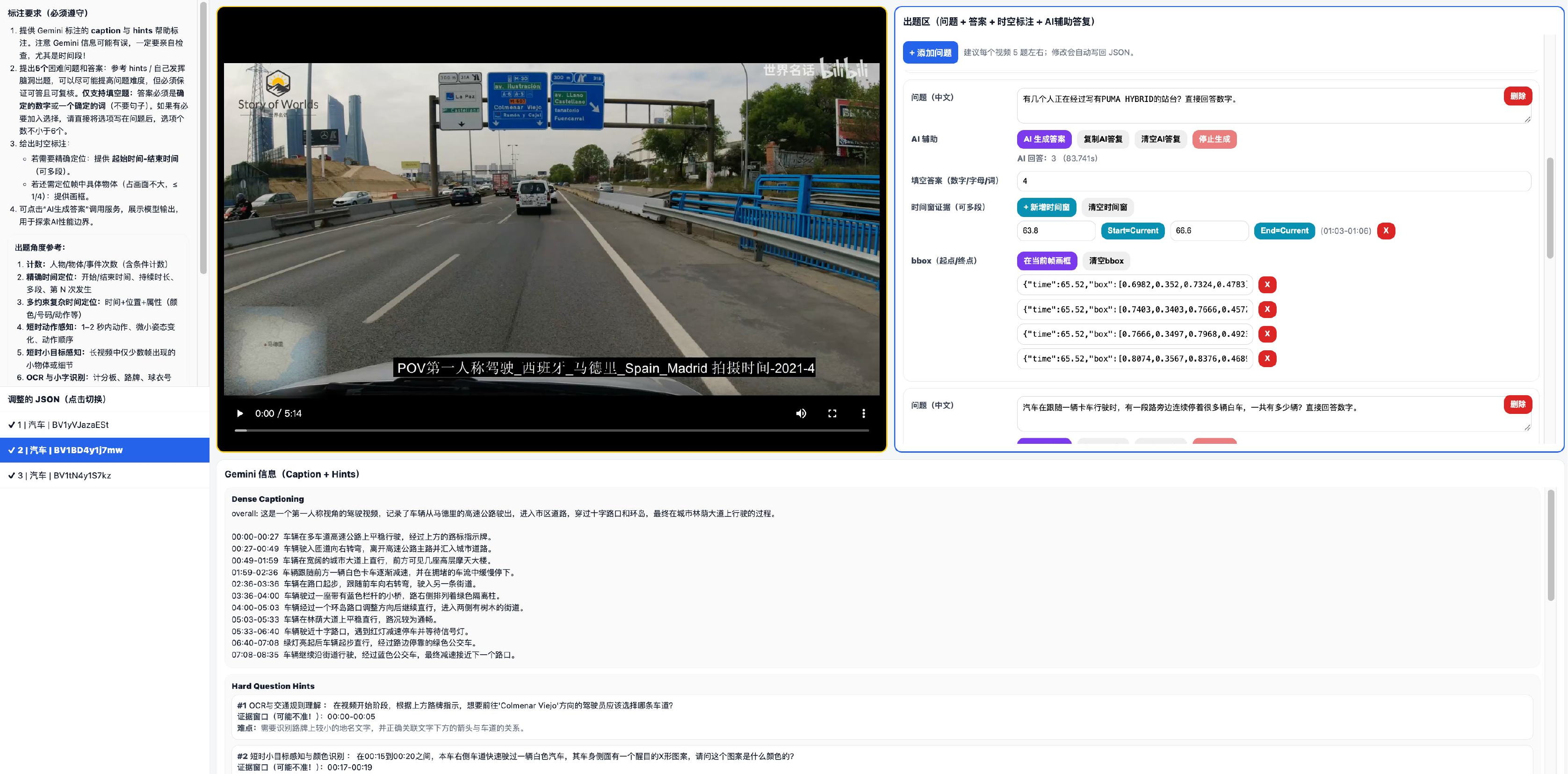}
    \caption{Annotation interface used for question construction and evidence labeling.}
    \label{fig:annotation_interface}
\end{figure}

\noindent
\textbf{Annotation Interface.}
As shown in Figure~\ref{fig:annotation_interface}, the human annotation interface includes the video player, Gemini-generated reference information, annotation guidelines, and input fields for questions, answers, and evidence.
The interface also supports direct API calls to Gemini models, allowing annotators to query the model during annotation.
This helps annotators understand the capability boundaries of video MLLMs and design challenging questions that expose potential weaknesses.

\begin{table}[t]
\centering
\small
\caption{Ablation on the frame sampling strategy. Results are reported as Level-3 accuracy.}
\setlength{\tabcolsep}{6pt}
\scalebox{0.96}{
\begin{tabular}{l|c|c|c|c|c}
\toprule
\textbf{Max Frames} & \textbf{48} & \textbf{96} & \textbf{192} & \textbf{384} & \textbf{768~(FPS=2.0)} \\
\midrule
Qwen3-VL-4B & 7.0 & \textbf{8.8} & 7.0 & 7.8 & 7.0 \\
Qwen3-VL-235B-A22B & 7.0 & \textbf{10.4} & 8.6 & 9.6 & 10.0 \\
\bottomrule
\end{tabular}
}
\label{tab:frame_ablation}
\end{table}

\FloatBarrier
\subsection{Human Performance Study}
\label{sec:ap_human}

The human study provides a reference for answering and grounding on a 50-question subset. Table~\ref{tab:main_results} reports Levels 3--5 and Table~\ref{tab:human_study} gives the Level-3 comparison on the same subset.

\begin{table}[t]
\centering
\small
\setlength{\tabcolsep}{2pt}
\caption{\textbf{Human performance on a 50-question subset of \name{}.}
The subset covers all 13 categories and includes 27 Chinese and 23 English questions. We report the overall Level-3 accuracy and results on several representative atomic capabilities.}
\label{tab:human_study}
\scalebox{0.9}{
\begin{tabular}{lccccccc}
\toprule
\textbf{Method}  & \textbf{Acc.} & \textbf{Count.} & \textbf{Small-Obj.}  & \textbf{Spatial Ori.} & \textbf{Obj.\ Track.} & \textbf{World Know.} & \textbf{Audio Perc.} \\
\midrule
Gemini-3-Pro   & 22.0 & 12.0 & 0.0  & 20.0 & 42.9  & 50.0 & \textbf{50.0}\\
Gemini-2.5-Pro  & 20.0 & 12.0 & 11.8  & 26.7 & 28.6 & 50.0 & 0.0 \\
Human      & \textbf{67.6} & \textbf{64.4} & \textbf{70.4} & \textbf{64.4} & \textbf{63.5} & \textbf{75.0} & 42.0 \\
\bottomrule
\end{tabular}
}
\end{table}

We further conduct a human performance study on a randomly sampled 50-question subset of \name{}.
The subset covers all 13 video categories and includes 27 Chinese questions and 23 English questions.
We recruit 20 participants whose birth years range from the 1970s to the 2000s, and all participants are independent of the dataset annotators.
Each participant answers at most 10 questions. Each question is answered by multiple human participants.
Human scores are averaged across participants within each question, then averaged across questions.

The Level-4/5 evaluation follows the same procedure, with participants providing temporal intervals and bounding boxes at designated key frames.
We apply the same IoU thresholds as in model evaluation.
Participants leave answers blank if they cannot find them in the video, and blank answers count as incorrect.
Human accuracy is 63.5\% at Level-4 and 55.8\% at Level-5.
Some errors involve imprecise intervals or boxes when questions require multiple evidence segments or targets.

The 50-question study provides a reference estimate under a practical evaluation budget.
Its scope is limited to this subset, where answering requires reviewing long videos, searching for subtle evidence, and replaying relevant segments.

Table~\ref{tab:human_study} shows that humans outperform frontier proprietary models on this subset.
The human accuracy reaches 67.6\%, while Gemini-3-Pro and Gemini-2.5-Pro achieve only 22.0\% and 20.0\%, respectively.
This large gap indicates that current video MLLMs remain far from human-level performance on this benchmark.

Humans consistently outperform the models on fine-grained perception tasks.
In particular, humans achieve much higher accuracy on \textit{Counting}, \textit{Small-Object}, \textit{Spatial Orientation}, and \textit{Object Tracking}.
These capabilities require identifying subtle visual cues and tracking objects across frames, which remain challenging for current video MLLMs.
Many questions require fine-grained inspection, so participants typically need around one hour to complete ten questions.

Compared with their performance on most other abilities, the Gemini models perform better on world-knowledge questions.
This suggests that the models possess substantial common knowledge about everyday scenarios.
For \textit{Audio Perception}, human performance is relatively low.
One possible reason is that participants may overlook audio cues while answering, or find it difficult to convert auditory information into precise textual responses.
Overall, the results suggest that the primary bottleneck of current video MLLMs lies in fine-grained visual perception in long videos.

\subsection{Evaluation Protocol Design}
\label{sec:ap_protocol}

Levels 4 and 5 require both correct answers and accurate evidence grounding.
Our protocol uses separate prompts for answer generation, temporal localization, and spatial grounding.
An alternative, joint prediction, requests the answer and required evidence in a single response.
We explain why we use separate prompts and compare the protocols to assess their impact on evaluation results.

\noindent
\textbf{Design rationale.}
The protocol evaluates temporal search and spatial inspection as separate tasks alongside answering.
For example, identifying the counseling-room scene and locating its empty chairs test different aspects of the evidence needed to answer the counting question (Figure~\ref{fig:cases}).

Manual box annotations are available at a few fixed key frames.
Because objects move or leave view, predictions must be compared with annotations at matching timestamps.
We provide the annotated timestamps and include these frames in the visual input, allowing spatial localization to be evaluated without requiring frame selection.
Evaluating arbitrary model-selected timestamps requires additional manual annotations, as in the joint-prediction experiment below.

Joint prediction requires stricter formats, such as JSON with separate fields for the answer, temporal intervals, and bounding boxes.
In our trials, some post-trained video reasoning specialists fail to follow this multi-field format.
We use separate prompts to limit each prediction's formatting demands.
Standalone tIoU and vIoU measure grounding regardless of answer correctness.

\noindent
\textbf{Relation to joint prediction.}
In joint prediction, the model receives only the video and question.
At Level-4, it predicts the answer and temporal intervals in a single response.
At Level-5, it predicts the answer, temporal intervals, and corresponding object bounding boxes with model-selected key-frame timestamps in a single response.

Our protocol reduces joint-generation demands through task-specific prompts and provides annotated key-frame timestamps for spatial grounding.
These relaxations motivate interpreting its scores as an optimistic estimate of joint answer-and-grounding performance.
Since MLLM predictions depend on the prompting conditions, this interpretation is heuristic and does not establish a theoretical upper bound.
The shared-subset results in Table~\ref{tab:joint_protocol} are consistent with this interpretation for the evaluated models.

\noindent
\textbf{Experimental verification.}
We compare the protocols on the same randomly sampled 100-question subset, with additional ground-truth boxes annotated at model-selected timestamps.
This annotation cost limits the comparison to a subset.
Joint prediction yields no accuracy improvement for any of the three models at either level (Table~\ref{tab:joint_protocol}).
Level-5 accuracy is 0--2\% with our protocol and 0--1\% with joint prediction.
Both settings show low answer-and-grounding accuracy for the three tested models on this subset.

\begin{table}[t]
\centering
\small
\caption{Separate versus joint prediction on a shared random 100-question subset (accuracy in \%).}
\label{tab:joint_protocol}
\begin{tabular}{lccccc}
\toprule
\multirow{2}{*}{Model} & \multirow{2}{*}{Level-3} & \multicolumn{2}{c}{Level-4} & \multicolumn{2}{c}{Level-5} \\
\cmidrule(lr){3-4}\cmidrule(lr){5-6}
& & Separate & Joint & Separate & Joint \\
\midrule
Gemini-3-Pro & 23.0 & 10.0 & 7.0 & 2.0 & 1.0 \\
Gemini-2.5-Pro & 21.0 & 9.0 & 7.0 & 2.0 & 0.0 \\
Qwen3-VL-4B & 9.0 & 3.0 & 2.0 & 0.0 & 0.0 \\
\bottomrule
\end{tabular}
\end{table}

\subsection{Robustness of Evaluation Results}
\label{sec:ap_reliability}

We examine whether variation across runs affects our main findings.
We evaluate five representative models five times at temperature 0.7, keeping the questions and other evaluation settings fixed.

Table~\ref{tab:repeat_runs} reports the mean accuracy and standard deviation for each model and level.
All reported standard deviations are below 0.7 percentage points.
These results characterize decoding variability on the fixed question set. They do not estimate uncertainty from sampling different questions or videos.

\begin{table}[t]
\centering
\small
\setlength{\tabcolsep}{4pt}
\caption{Accuracy (\%, mean $\pm$ standard deviation) over five runs at temperature 0.7.}
\label{tab:repeat_runs}
\begin{tabular}{lccccc}
\toprule
Model & Level-1 & Level-2 & Level-3 & Level-4 & Level-5 \\
\midrule
Gemini-3-Pro & $23.6\pm0.36$ & $20.2\pm0.50$ & $16.9\pm0.30$ & $7.8\pm0.35$ & $0.9\pm0.09$ \\
Gemini-2.5-Pro & $25.3\pm0.48$ & $21.0\pm0.35$ & $16.4\pm0.32$ & $7.3\pm0.22$ & $0.6\pm0.09$ \\
Qwen3-VL-4B & $22.4\pm0.52$ & $16.5\pm0.58$ & $7.7\pm0.31$ & $1.2\pm0.15$ & $0.0\pm0.00$ \\
% Qwen2.5-VL-7B L4 runs (%): 0, 0, 0, 0, 0.2. Mean = 0.04; sample SD (ddof=1) = 0.08944.
Qwen2.5-VL-7B & $16.1\pm0.44$ & $12.4\pm0.50$ & $5.8\pm0.33$ & $0.0\pm0.09$ & $0.0\pm0.00$ \\
VideoChat-R1.5-7B & $17.6\pm0.50$ & $12.2\pm0.68$ & $8.2\pm0.42$ & $0.0\pm0.09$ & $0.0\pm0.00$ \\
\bottomrule
\end{tabular}
\end{table}

\begin{figure}[t]
    \centering
    \includegraphics[width=0.85\linewidth,height=0.80\textheight,keepaspectratio]{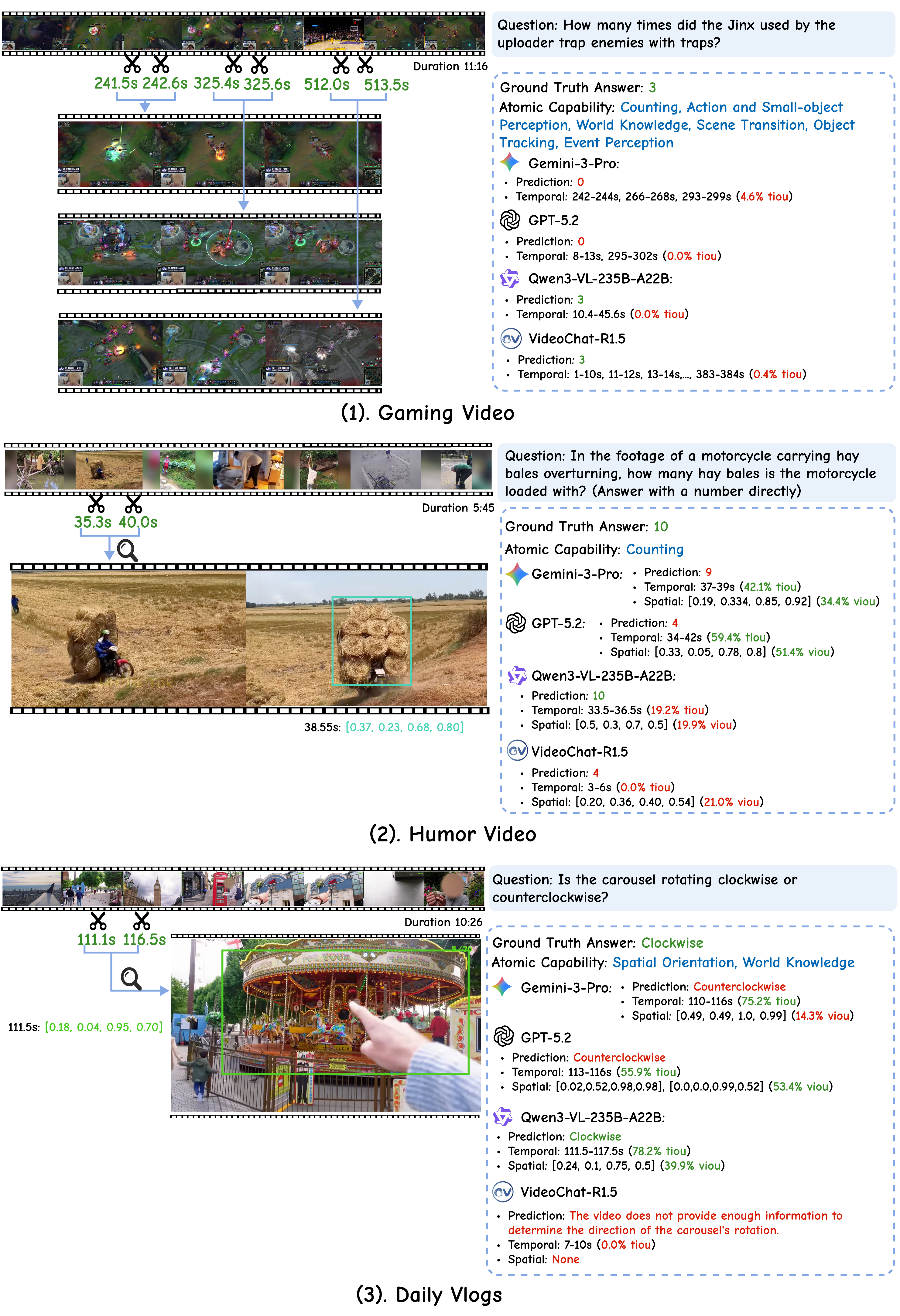}
    \caption{Examples from \textbf{\name{}} organized by video category, with predictions from several models. The left side shows the annotated spatio-temporal evidence. In the text on the right, green indicates correct answers or IoU $>$ 0.3, while red indicates incorrect answers or IoU $\le$ 0.3. Spatial coordinates are normalized to the $[0,1]$ range for key timestamps. \textbf{(1)} The model answers correctly but fails to locate the correct temporal evidence. \textbf{(2)(3)} Gemini and GPT show relatively accurate localization, but still make mistakes in counting or spatial orientation. Qwen3-VL-235B achieves Level-5 in (3). }
    \label{fig:vis_01}
\end{figure}

\begin{figure}[t]
    \centering
    \includegraphics[width=0.96\linewidth,height=0.80\textheight,keepaspectratio]{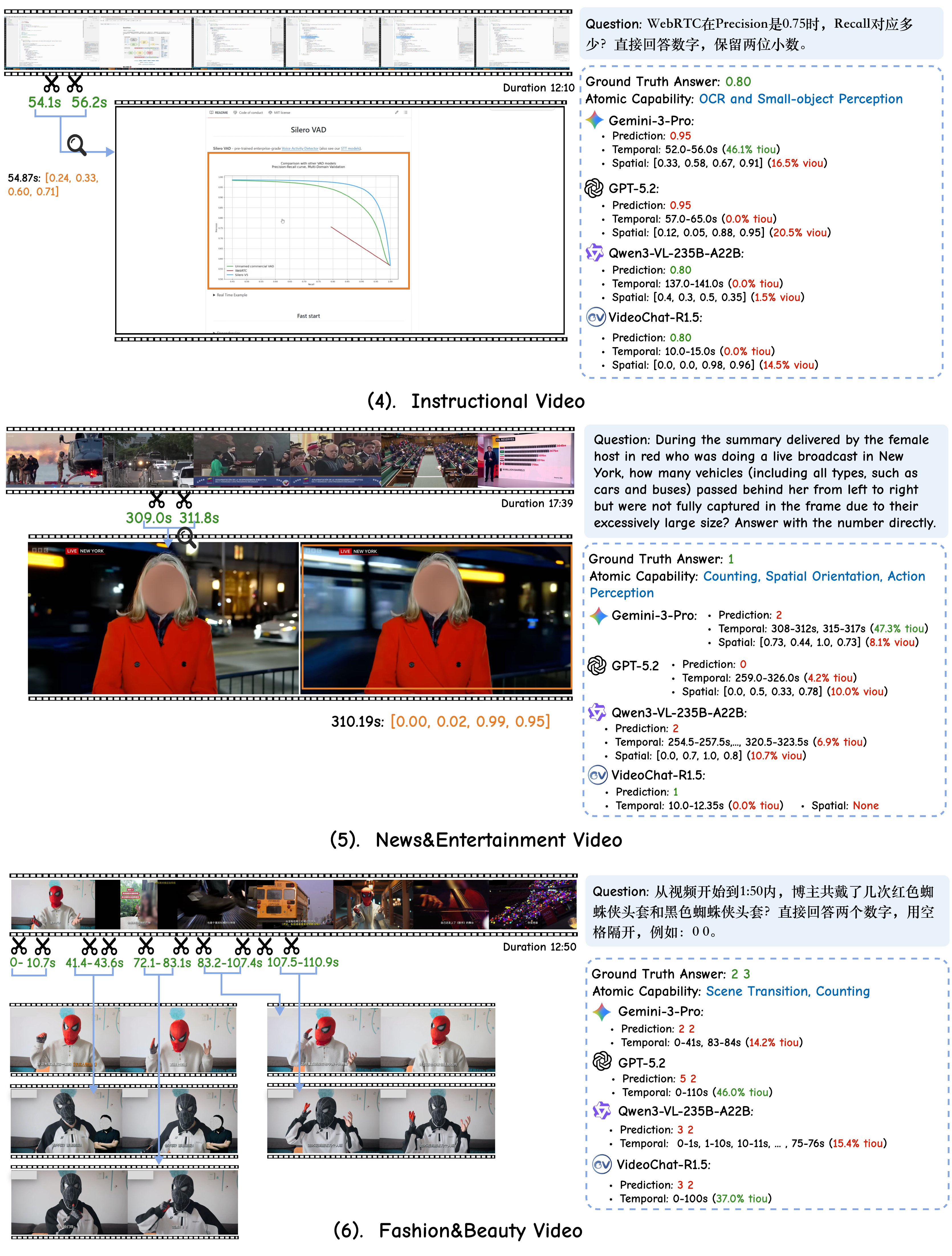}
    \caption{Examples from \textbf{\name{}} by video category, with model predictions. \textbf{(4)} Models struggle to localize and interpret chart information, even when the final answer is correct. \textbf{(5)(6)} Models perform poorly on counting tasks under complex constraints with easily confused objects. }
    \label{fig:vis_02}
\end{figure}

\begin{figure}[t]
    \centering
    \includegraphics[width=0.96\linewidth,height=0.80\textheight,keepaspectratio]{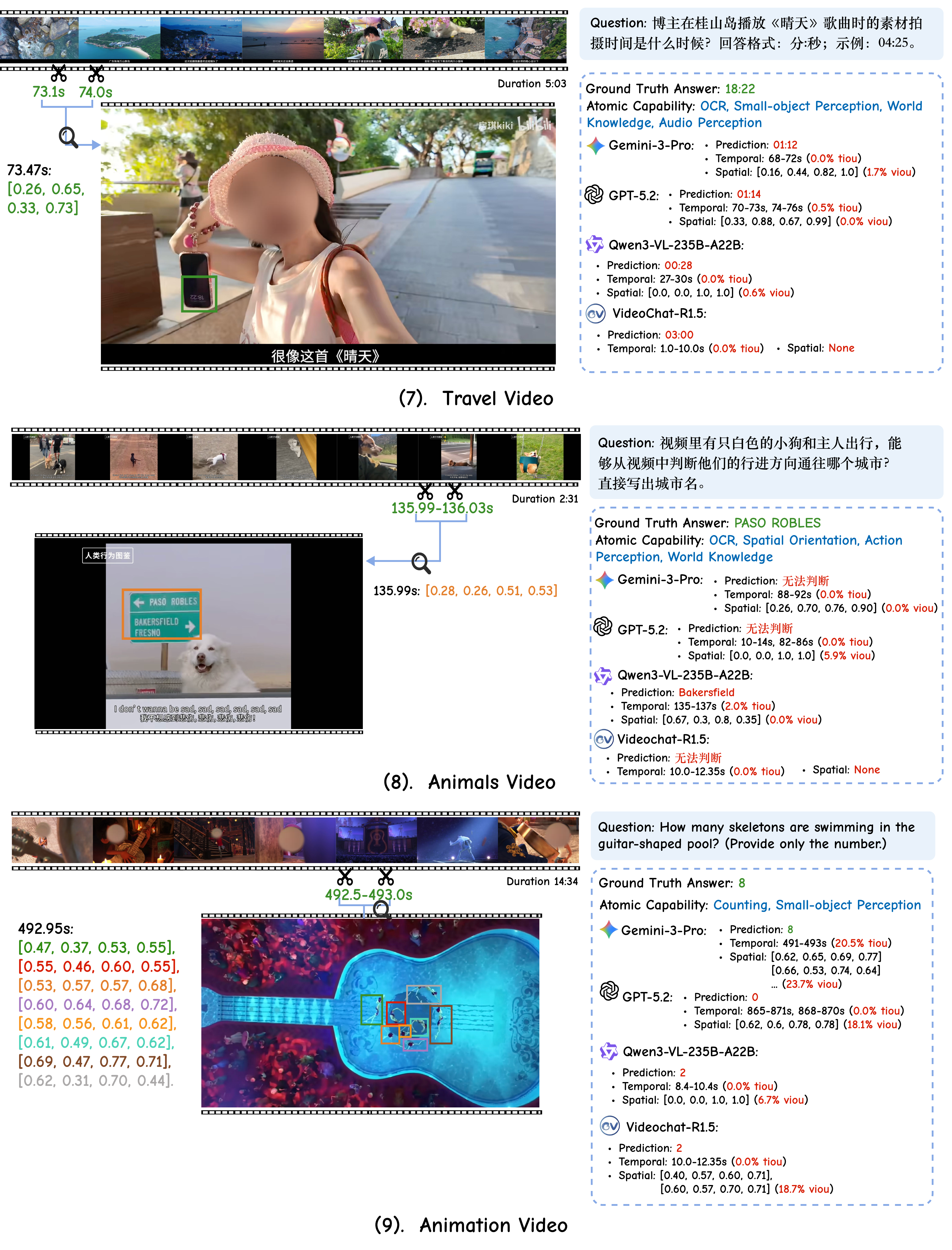}
    \caption{Examples from \textbf{\name{}} with model predictions. \textbf{(7)} Models struggle with a small visual cue in an audio-conditioned question. \textbf{(8)} The annotated event lasts 0.04s. A 135--137s prediction covers it but receives only 2.0\% tIoU, illustrating sensitivity to temporal precision. \textbf{(9)} Gemini answers correctly, but its wider temporal window and spatial predictions fall below the grounding thresholds.}
    \label{fig:vis_03}
\end{figure}

\subsection{Additional Visualizations and Error Cases}
\label{sec:ap_vis}

To better illustrate typical reasoning failures of current video MLLMs, we present several representative examples from \name{} in Figures~\ref{fig:vis_01},~\ref{fig:vis_02}, and~\ref{fig:vis_03}.
These cases show errors in temporal and spatial grounding, including incorrect evidence predictions accompanying correct answers.
They also illustrate weaknesses in counting, small-object perception, and spatial orientation.

\subsection{Related Work}
\label{sec:ap_related_works}

% \section{Related Works}
% \label{sec:related_works}

\noindent
\textbf{Video MLLM Benchmarks.}
Most video MLLM benchmarks~\citep{VideoMME,VideoMMMU,LongVideoBench,wang2025lvbench,MVBench,ALLVB,MMVU,MMBenchVideo,TempCompass} evaluate models' ability to understand dynamic scenes, integrate multimodal signals, and reason over long videos.
Video-MME~\citep{VideoMME} provides a comprehensive evaluation suite across short, medium, and long videos.
MVBench~\citep{MVBench} focuses on temporally sensitive tasks, such as action understanding, object interaction, and event reasoning.
MMBench-Video~\citep{MMBenchVideo} extends multimodal benchmark design to video scenarios with broad coverage of perception and reasoning skills.
LVBench~\citep{wang2025lvbench} and LongVideoBench~\citep{LongVideoBench} stress long-form video comprehension, where key information may be distributed across distant temporal positions.
Video-MMMU~\citep{VideoMMMU} targets knowledge-intensive instructional videos and evaluates whether models can connect visual evidence with domain concepts.
CG-Bench~\citep{CGBench} introduces clue-grounded question answering and tests whether models can localize supporting temporal clues.
ALLVB~\citep{ALLVB} and MMVU~\citep{MMVU} broaden evaluation toward diverse video categories and university-level multimodal understanding.
TempCompass~\citep{TempCompass} focuses on temporal reasoning, including event order, duration, frequency, and temporal relation understanding.
These benchmarks have greatly advanced video MLLM evaluation.
However, many of them rely on multiple-choice protocols or coarse answer accuracy, which may overestimate model capability.
They also rarely verify whether a correct answer is supported by precise spatio-temporal evidence.
In contrast, \name{} is designed to probe the performance boundary of current video MLLMs through substantially harder open-ended questions and a hierarchical evaluation protocol that exposes unresolved challenges in answering, temporal grounding, and spatial grounding.

\noindent
\textbf{Detailed Spatial Benchmarks.}
Another line of work studies fine-grained visual perception and deeper reasoning grounded in visual evidence.
In the image domain, ZeroBench~\citep{roberts2025zerobench} constructs extremely difficult visual questions that require careful inspection of subtle image details.
BLINK~\citep{fu2024blink} evaluates perceptual abilities that are easy for humans but difficult for MLLMs, such as relative depth, counting, visual correspondence, and spatial relations.
TreeBench~\citep{wang2025traceable} emphasizes traceable visual reasoning, where models must follow structured evidence paths rather than relying on shortcut recognition.
These image benchmarks reveal that strong global semantics do not imply reliable detailed perception.
In videos, related efforts mainly fall into two directions: spatio-temporal localization and spatial intelligence.
V-STaR~\citep{VSTaR}, Know-Show~\citep{sugandhika2025know}, and ToG-Bench~\citep{xu2025tog} evaluate object and event grounding through spatio-temporal reasoning, action-conditioned evidence, and embodied tasks.
Dr.V~\citep{luo2025dr} uses spatio-temporal grounding to diagnose video hallucination on examples derived from existing public datasets.
In contrast, \name{} uses newly collected long videos with fully human-annotated questions, answers, temporal intervals, and spatial boxes.
VidSTG~\citep{vidstg} and HC-STVG~\citep{tang2021hcstvg} provide classic settings for spatio-temporal video grounding, where models localize queried objects or events in space and time.
However, they focus on grounding given queries rather than solving hard QA problems.
VSI-Bench~\citep{yang2025vsibench}, SpatialTree~\citep{xiao2025spatialtree}, and MMSI-Video-Bench~\citep{lin2025mmsi} instead examine spatial relations, motion direction, and structural understanding in dynamic scenes.
However, they mainly evaluate spatial intelligence itself, without requiring joint QA and evidence localization.

\noindent
\textbf{Thinking-with-Videos.}
Recent multimodal models such as Gemini-3~\citep{gemini-3-pro}, GPT-5.2~\citep{gpt-5.2}, Qwen3-VL~\citep{bai2025qwen3vl}, and post-trained variants like Video-R1~\citep{feng2025video-r1} already demonstrate structured reasoning and improved temporal understanding.
Inspired by OpenAI-o3~\citep{openai-o3}, a growing line of work~\citep{yan2025videochat,ding2025videozoomer,zhang2025rewatch,ouyang2025conan,meng2025open,wang2025pixel,zeng2026videoo3,zhang2025thinking,zhang2025rewatch,wang2025video-thinker,fu2025love,yuan2025videoexplorerthinkvideosagentic,ge2025framemind,yang2025longvt,tang2025video,gao20251+} studies thinking-with-videos, where reasoning is interleaved with explicit visual grounding.
VideoChat-R1.5~\citep{yan2025videochat}, VideoZoomer~\citep{ding2025videozoomer}, and Video-o3~\citep{zeng2026videoo3} focus on temporal evidence seeking via iterative zoom-in to refine relevant intervals.
Open-o3-Video~\citep{meng2025open} directly emits timestamps and bounding boxes within a reasoning trace without external tool calls.
These approaches make evidence more explicit and motivate evaluation on difficult long-video questions with fine-grained manual evidence annotations.
Our benchmark is designed to fill this gap by evaluating challenging questions and explicitly verifying temporal and spatial evidence.

\subsection{Limitations and Future Work}
\label{sec:ap_limitations}

\name{} focuses on deliberately challenging questions from a limited collection of videos.
Its results characterize this diagnostic set. Broader domains, languages, and video sources are needed to assess generalization.
Evidence annotations consist of temporal intervals and sparse key-frame boxes.
Extending coverage and adding trajectories or dense masks would support finer-grained evaluation.
These extensions can also help assess thinking-with-videos systems that actively search, zoom, and verify evidence before answering.

\end{document}